\documentclass{article}

\usepackage{PRIMEarxiv}

\usepackage[utf8]{inputenc} 
\usepackage[T1]{fontenc}    
\usepackage[hidelinks]{hyperref}       
\usepackage{url}            
\usepackage{booktabs}       
\usepackage{amsfonts}       
\usepackage{nicefrac}       
\usepackage{microtype}      
\usepackage{lipsum}
\usepackage{fancyhdr}       
\usepackage{graphicx}       
\usepackage{multirow}
\usepackage{xcolor}
\usepackage{soul}
\usepackage{amsmath}
\usepackage{siunitx}
\usepackage{comment}
\usepackage{textcomp, gensymb} 
\usepackage{placeins}
\usepackage{makecell}
\usepackage[inline]{enumitem}
\usepackage{caption}
\usepackage{ragged2e}\usepackage[numbers]{natbib}
\DeclareCaptionLabelFormat{AppendixTables}{Table A.#2}

\graphicspath{{media/}}     

\title{ \textcolor{black}{FIESTA: Autoencoders for accurate fiber segmentation in tractography}}\vspace{-0.cm}

\author{
  Félix Dumais \\
  Sherbrooke Connectivity Imaging Lab (SCIL) \\
  Videos \& Images Theory and Analytics Lab (VITAL) \\
  Department of Computer Science\\
  Université de Sherbrooke, Canada\\
  \texttt{felix.dumais@usherbrooke.ca} \\
  \And
  Jon Haitz Legarreta\textcolor{black}{*} \\
  \textcolor{black}{Department of Radiology} \\
  \textcolor{black}{Brigham and Women's Hospital} \\
  \textcolor{black}{Mass General Brigham/Harvard Medical School, USA} \\
  \And
  Carl Lemaire \\
  Centre de Calcul Scientifique\\
  Université de Sherbrooke, Canada\\
  \And
  Philippe Poulin \\
  Sherbrooke Connectivity Imaging Lab (SCIL) \\
  Videos \& Images Theory and Analytics Lab (VITAL) \\
  Department of Computer Science\\
  Université de Sherbrooke, Canada\\
  \And
  François Rheault \\
  Medical Imaging and Neuroinformatic (MINi) Lab \\
  Department of Computer Science\\
  Université de Sherbrooke, Canada\\
  \And
  Laurent Petit \\
  Groupe d'Imagerie Neurofonctionnelle (GIN) \\
  CNRS, CEA, IMN, GIN, UMR 5293, F-33000 Bordeaux \\
  Université de Bordeaux, France\\
  \And
  \textcolor{black}{Muhamed Barakovic} \\
  \textcolor{black}{Pharma Research and Early Development} \\
  \textcolor{black}{Neuroscience and Rare Diseases} \\
  \textcolor{black}{Roche Innovation Center Basel} \\
  \textcolor{black}{F. Hoffmann-La Roche Ltd., Basel, Switzerland}\\
  \And
  \textcolor{black}{Stefano Magon} \\
  \textcolor{black}{Pharma Research and Early Development} \\
  \textcolor{black}{Neuroscience and Rare Diseases} \\
  \textcolor{black}{Roche Innovation Center Basel} \\
  \textcolor{black}{F. Hoffmann-La Roche Ltd., Basel, Switzerland}\\
  \And
  Maxime Descoteaux*\textcolor{black}{*} \\
  Sherbrooke Connectivity Imaging Lab (SCIL) \\
  Department of Computer Science\\
  Université de Sherbrooke, Canada\\
  Imeka Solutions inc, Sherbrooke, Canada\\
  \And
  Pierre-Marc Jodoin*\textcolor{black}{*} \\
  Videos \& Images Theory and Analytics Lab (VITAL) \\
  Department of Computer Science\\
  Université de Sherbrooke, Canada\\
  Imeka Solutions inc, Sherbrooke, Canada\\
  \AND
  \\\textcolor{black}{\textbf{for the Alzheimer's Disease Neuroimaging Initiative***}}\\
  \\
  }

\begin{document}
\maketitle

\begingroup
\leftskip=1cm plus 0.5fil \rightskip=1cm plus -0.5fil
\textcolor{black}{*Work done while at the SCIL and VITAL labs, Department of Computer Science, Université de Sherbrooke, Canada}

*\textcolor{black}{*}Co-senior author. These authors contributed equally.

\textcolor{black}{***Data used in preparation of this article were obtained from the Alzheimer's Disease Neuroimaging Initiative (ADNI) database (adni.loni.usc.edu). As such, the investigators within the ADNI contributed to the design and implementation of ADNI and/or provided data, but did not participate in analysis or writing of this report. A complete listing of ADNI investigators can be found at: \url{http://adni.loni.usc.edu/wp-content/uploads/how_to_apply/ADNI_Acknowledgement_List.pdf}}

\textcolor{black}{\textit{Note}: Conflicts of Interest - Maxime Descoteaux and Pierre-Marc Jodoin report membership and employment with Imeka Solutions inc. Patent \#17/337,413 is pending to Imeka Solutions inc. with inventors Jon Haitz Legarreta, Maxime Descoteaux and Pierre-Marc Jodoin. Muhamed Barakovic is an employee of Hays plc and a consultant for F. Hoffmann-La Roche Ltd. Stefano Magon is an employee and shareholder of F. Hoffmann-La Roche Ltd.}

\par
\endgroup

\pagebreak
\vspace{-1cm}

\begin{abstract}
White matter bundle segmentation is a cornerstone of modern tractography to study the brain's structural connectivity in domains such as neurological disorders, neurosurgery, and aging. In this study, we present FIESTA (\textcolor{black}{\textit{FIbEr Segmentation in Tractography using Autoencoders}}), a reliable and robust, fully automated, and easily semi-automatically calibrated pipeline based on deep autoencoders that can dissect and fully populate \textcolor{black}{white matter} bundles. This pipeline is built upon \textcolor{black}{previous works} that demonstrated how autoencoders can be used successfully for streamline filtering, \textcolor{black}{bundle segmentation}, and streamline generation in tractography. Our proposed method improves \textcolor{black}{bundle segmentation} coverage by recovering hard-to-track bundles with generative sampling through the latent space seeding of the subject bundle and the atlas bundle. A latent space of streamlines is learned using autoencoder-based modeling combined with contrastive learning. Using an atlas of bundles in standard space (MNI), our proposed method segments new tractograms using the autoencoder latent distance between each tractogram streamline and its closest neighbor bundle in the atlas of bundles. Intra-subject bundle reliability is improved by recovering hard-to-track streamlines, using the autoencoder to generate new streamlines that increase the spatial coverage of each bundle while remaining anatomically \textcolor{black}{correct}. Results show that our method is more reliable than state-of-the-art automated virtual dissection methods such as {\em RecoBundles}, {\em RecoBundlesX}, {\em TractSeg}, {\em White Matter Analysis} and {\em XTRACT}. \textcolor{black}{Our framework allows for the transition from one anatomical bundle definition to another with marginal calibration \textcolor{black}{efforts}}. Overall, these results show that our framework improves the practicality and usability of current state-of-the-art \textcolor{black}{bundle segmentation} framework
\end{abstract}

\vspace{-0.2cm}
\vspace{-0.3cm}

\section{Introduction}
\label{sec:introduction}
\vspace{-0.2cm}
White matter (WM) fiber tractography is a well-established method for brain connectivity analysis. It is currently the only non-invasive method able to investigate brain WM pathways \textit{in vivo}. By using the local water diffusion information from diffusion-weighted Magnetic Resonance Imaging (dMRI) images, one can infer the local orientation of the underlying WM streamlines \cite{descoteaux_deterministic_2009, descoteaux_regularized_2007, tournier_robust_2007} and use it to numerically reconstruct WM pathways. Over the years, many challenges have been tackled to improve this modeling technique such as \textit{Global Tractography} \cite{kreher_gibbs_2008, mangin_toward_2013, christiaens_global_2015}, \textit{Probabilistic Tractography} \cite{descoteaux_deterministic_2009, tournier_diffusion_2011, tournier_mrtrix_2012}, \textit{Particle Filtering Tractography} (PFT) \cite{girard_towards_2014}, \textit{Bundle-Specific Tractography} \cite{wasserthal_tract_2018, rheault_bundle-specific_2019}, or \textit{Surface-Enhanced Tractography} (SET) \cite{st-onge_surface-enhanced_2018}.

The usability of WM tractography often comes from one's ability to filter and group streamlines into WM bundles. In this work, we refer to \textbf{filtering} as any method able to remove implausible streamlines from whole-brain tractograms (filtering methods yield implausible-free whole-brain tractograms). \textcolor{black}{Filtering will later be discussed as it is used inside the \textcolor{black}{bundle segmentation} module to filter and segment tractograms (c.f. Fig. \ref{fig:ae_binta}b and section \ref{sec:tractogram_filtering_and_clustering}).} Furthermore, there exists automated grouping methods that are 100\% data-driven, fully unsupervised, whilst other methods follow pre-defined anatomical definitions. Thus, unsupervised grouping methods will be referred to as \textbf{clustering} (clustering methods yield streamline clusters), whilst grouping methods following pre-defined anatomical definitions will be referred to as \textbf{segmenting} (segmenting methods yield streamline bundles).  
The most faithful way to extract WM bundles from a tractogram is by asking a neuroanatomist to manually dissect bundles of interest (BOI). Manual dissection of bundles is long and tedious, and is prone to large inter- and intra-expert variability \cite{rheault_tractostorm_2020, rheault_tractostorm_2022}. Automated methods, such as \textit{QuickBundles} (QB) \cite{garyfallidis_quickbundles_2012}, \textit{QuickBundlesX} (QBx) \cite{garyfallidis_quickbundlesx_2015}, \textit{Deep Fiber Clustering} (DFC) \cite{chen_deep_2023}, \textit{RecoBundles} (RB) \cite{garyfallidis_recognition_2018}, \textit{RecoBundlesX} (RBx) \cite{rheault_analyse_2020}, \textit{TractSeg} \cite{wasserthal_tractseg_2018, wasserthal_combined_2019, wasserthal_multiparametric_2020}, \textit{XTRACT} \cite{warrington_xtract_2020}, and \textit{White Matter Analysis} (WMA) \cite{odonnell_automatic_2007, odonnell_unbiased_2012, zhang_anatomically_2018}, amongst others, have been proposed to accelerate and increase the reproducibility of this process.

Unfortunately, these grouping methods are not void of limitations. First, clustering methods are fully unsupervised, thus excluding prior knowledge of each WM bundle's class membership (\textit{QuickBundles}, \textit{QuickBundlesX}, DFC). Second, even if there is  a lack of consensus in the community over each bundle's anatomical definition \cite{rheault_common_2020}, some state-of-the-art methods do not allow an easy modification of anatomical definitions (\textit{TractSeg}, \textit{XTRACT}). In fact, if a change in bundle definitions is needed with \textit{TractSeg}~\cite{wasserthal_tractseg_2018}, a full reannotation of the training dataset is required alongside with a complete retraining of its three neural networks. 
Contrastingly, \textit{XTRACT} \cite{warrington_xtract_2020} heavily relies on regions of interest (ROI) drawn by expert neuroanatomists to work properly. Those ROIs are usually hard to get and prone to inter-expert variability. Third, methods such as WMA are impractical in a large scale context because they rely on an affinity matrix computed using the pairwise distance of each streamline in a whole-brain tractogram. Such methods have prohibitive memory usage when the tractogram has more than \num[group-separator={,}]{500000} streamlines \cite{wasserthal_tractseg_2018}. Fourth, methods such as \textit{RecoBundles} \cite{garyfallidis_recognition_2018} and \textit{RecoBundlesX} \cite{rheault_analyse_2020} require the non-trivial calibration of several parameters and are more or less reliable in a test-retest analysis (see \textit{RecoBundles} and \textit{RecoBundlesX} results in section \ref{sec:results}). Finally, knowing all \textcolor{black}{possible} limitations of dMRI tractography \cite{rheault_common_2020}, \textcolor{black}{segmentation} and clustering methods are tied to the tracking algorithms' ability to recover hard-to-track WM bundles in the first place.


In this paper, we present FIESTA (\textcolor{black}{\textit{FIbEr Segmentation in Tractography using Autoencoders}}), a reliable and robust, fully automated, and easily semi-automatically calibrated pipeline based on deep autoencoders that can dissect and fully populate WM bundles. \textcolor{black}{Thus, FIESTA is a \textbf{filtering} and \textbf{\textcolor{black}{bundle segmentation}} (not \textbf{clustering}) pipeline whose output is subsequently improved by  an autoencoder-based generative streamline sampling method.} FIESTA allows an easy change in its bundle definitions, depending on the need, with a marginal calibrating time. This pipeline is built upon FINTA, CINTA, and GESTA methods \cite{legarreta_filtering_2021, Legarreta:MICCAI-CDMRI:2022, legarreta_generative_2022} that demonstrated how autoencoders can be used successfully for filtering, \textcolor{black}{bundle segmentation}, and streamline generation in tractography. 

\subsection{Related work}
\label{sec:related_work}
Over the years, many methods have been developed to ease the interpretability of whole-brain tractograms, such as filtering methods \cite{legarreta_filtering_2021, maier-hein_challenge_2017, petit_half_2021, jorgens_challenges_2021, sotiropoulos_building_2019} and clustering methods \cite{garyfallidis_quickbundles_2012, garyfallidis_quickbundlesx_2015, chen_deep_2021, chen_dfc_2022, odonnell_automatic_2007, Wang_2011, Visser_2011}. Unfortunately, filtering methods do not allow to easily target anatomical regions as tractograms are not grouped in streamline bundles. On the other hand, streamline clustering algorithms, a class of methods designed to group streamlines with similar properties, are typically built upon unsupervised machine learning approaches such as mixture models, spectral clustering and hierarchical clustering. Unfortunately, clustering methods suffer from a major drawback: they offer no control to which cluster belongs to which WM anatomical region. 

Therefore, \textcolor{black}{bundle segmentation} comes as a solution for this problem. It aims to give streamlines an anatomical label. 
Many methods have been proposed in the literature to bundle whole-brain tractograms, such as TRACULA \cite{yendiki_automated_2011}, \textit{TractQuerier} \cite{wassermann_white_2016}, or \textit{Classifyber} \cite{berto_classifyber_2021}. In this work, we will confine our reference methods to \textit{RecoBundles} \cite{garyfallidis_recognition_2018} and \textit{RecoBundlesX} \cite{rheault_analyse_2020}, WMA \cite{odonnell_automatic_2007, odonnell_unbiased_2012, zhang_anatomically_2018}, \textit{TractSeg} \cite{wasserthal_tract_2018, wasserthal_tractseg_2018, wasserthal_combined_2019, wasserthal_multiparametric_2020} and \textit{XTRACT} \cite{warrington_xtract_2020}. Reference methods for this work are based on different criteria. The main goal is to capture the representativeness of the different existing state-of-the-art methods. \textit{RecoBundles} and \textit{RecoBundlesX} were selected for their ease of implementation, and because author F.R. developed the latter. \textit{XTRACT} was selected to represent ROI-based \textcolor{black}{bundle segmentation} methods. WMA was selected to represent methods based on a non-physical embedding space. Finally, \textit{TractSeg} was selected to capture the power of supervised deep learning methods. 

\textit{RecoBundles} is a well-established method based upon atlas bundles pre-segmented by medical experts.
When a new tractogram needs to be bundled, a streamline-based linear registration \cite{garyfallidis_robust_2015} is done between the target and the atlas tractograms, bringing the tractogram and the atlas in the same space.
After this co-registration, a sequence of operations is performed for each bundle in the atlas. First, the target tractogram is pruned to remove irrelevant streamlines to the current bundle of interest. Then, a local streamline-based linear registration is performed to better match the target anatomy. Finally, a second pruning operation is performed using a stricter distance threshold to isolate the streamlines that have a high shape similarity to the current bundle of interest in the atlas. All the aforementioned steps are performed using centroids rather than individual streamlines to speed up processing. 

\textit{RecoBundlesX} is an improved version of \textit{RecoBundles} using many iterations of \textit{RecoBundles} over multiple atlases with different sets of parameters. Varying parameters are \begin{enumerate*}[label=(\roman*)] \item whole-brain tractograms \textit{QuickBundles} clustering thresholds, \item atlas bundle \textit{QuickBundles} clustering thresholds, and \item pruning thresholds \end{enumerate*}. While being much more reliable than \textit{RecoBundles}, this method is time-consuming since \textit{RecoBundles} is launched many times.

\textit{White Matter Analysis} (WMA), proposed by \citet{odonnell_automatic_2007}, is an atlas-based bundle recognition method where streamline classification is done in an embedding space built with spectral clustering \cite{odonnell_method_2006, hutchison_white_2005}. The atlas streamlines are represented as a point in the embedding space, where each point is assigned to a bundle class. New tractogram streamlines are bundled using the closest atlas centroid as the most probable class. One drawback of the method is that it requires computing a $N\times N$ pairwise streamline distance affinity matrix, which is computationally prohibitive for a reasonably sized tractogram. Therefore, the Nystrom approximation \cite{fowlkes_spectral_2004} is done as a trade-off to approximate the affinity matrix.

\textcolor{black}{ROI-based methods, such as }\cite{Zhang:Neuroimage:2010}\textcolor{black}{, including the White Matter Query Language (WMQL) }\cite{wassermann_white_2016}, use a pre-defined atlas of ROIs \textcolor{black}{to accomplish the bundle segmentation task.} ROIs are defined in a standard space and contain starting areas, inclusion and exclusion zones, and termination areas for each bundle. The regions are non-linearly registered to the diffusion space and used to bundle raw tractograms. \textcolor{black}{More recently, }\citet{warrington_xtract_2020} \textcolor{black}{proposed the \textit{XTRACT} protocol, which leverages ROI-based methods in order to achieve an improved generalizability.}

\textit{TractSeg} is a state-of-the-art deep learning \textcolor{black}{bundle segmentation} method. Presented over many articles \cite{wasserthal_tract_2018, wasserthal_tractseg_2018, wasserthal_combined_2019, wasserthal_multiparametric_2020}, the latest version works in three steps. First, a constrained spherical deconvolution (CSD) is used over the diffusion orientation distribution function (dODF) to extract the three principal peaks from the obtained fiber orientation distribution function (fODF). Next, peak maps are given to three U-Net neural networks \cite{ronneberger_miccai_u-net_2015} yielding $B=72$ bundle segmentation maps, 2 $\times$ $B$ start and end bundle segmentation maps, and $B$ bundle-specific \textit{Tract Orientation Maps} (TOM). Finally, probabilistic tractography is done over each TOM within their respective bundle segmentation maps and making sure start and end regions are respected for each streamline. 

\subsection{Contributions}
\label{sec:contributions}
In this paper, we present a semi-supervised \textcolor{black}{bundle segmentation} method called FIESTA. This work is the natural extension of FINTA \cite{legarreta_filtering_2021}, CINTA \cite{Legarreta:MICCAI-CDMRI:2022}, and GESTA \cite{legarreta_generative_2022}, which use a deep convolutional autoencoder to project streamlines into a lower, more structured, smoother and locally linear dimensional space to either filter, segment, or generate WM streamlines (\textcolor{black}{c.f. Supplemental section A.6 for more details}). To better structure the latent representation, we trained the autoencoder with a contrastive loss \cite{hadsell_dimensionality_2006} informed by \textit{QuickBundlesX} clustering.  FIESTA uses an autoencoder to filter and bundle whole-brain tractograms based on a given bundle atlas. It also takes advantage of its latent space sampling strategy to synthesize new streamlines and improve the coverage of its bundles. To avoid confusion and to be sure to distinguish the slight method variations between previous works and the current implemented methods, we renamed the implemented 2-in-1 filtering and \textcolor{black}{bundle segmentation} method as \textcolor{black}{FINTA-multibundle}, and the implemented generative method as \textcolor{black}{GESTA-gmm} \textcolor{black}{(Gaussian mixture model)} (c.f. section \ref{sec:material_and_methods}). This work makes the following five contributions:

\begin{enumerate}
    \item FIESTA leverages the power of previous fiber autoencoders \cite{legarreta_filtering_2021, Legarreta:MICCAI-CDMRI:2022, legarreta_generative_2022} and overcomes their individual limitations, yielding a pipeline able to work in an end-to-end \textit{in vivo} dMRI data analysis scenario;
    \item FIESTA is more reliable than current state-of-the-art automatic \textcolor{black}{bundle segmentation} methods;
    \item We show that the usage of generative sampling improves the bundle-wise volumetric reliability;
    \item We show that we can train and use contrastive learning based on \textit{QuickBundlesX} clusters to build a useful latent representation of a whole-brain tractogram;
    \item Bundle definitions are easily editable in FIESTA without the need to re-train a neural network.
\end{enumerate}


\section{Methods}
\label{sec:material_and_methods}

\begin{figure}[!ht]
    \centering
    \includegraphics[width=1\textwidth]{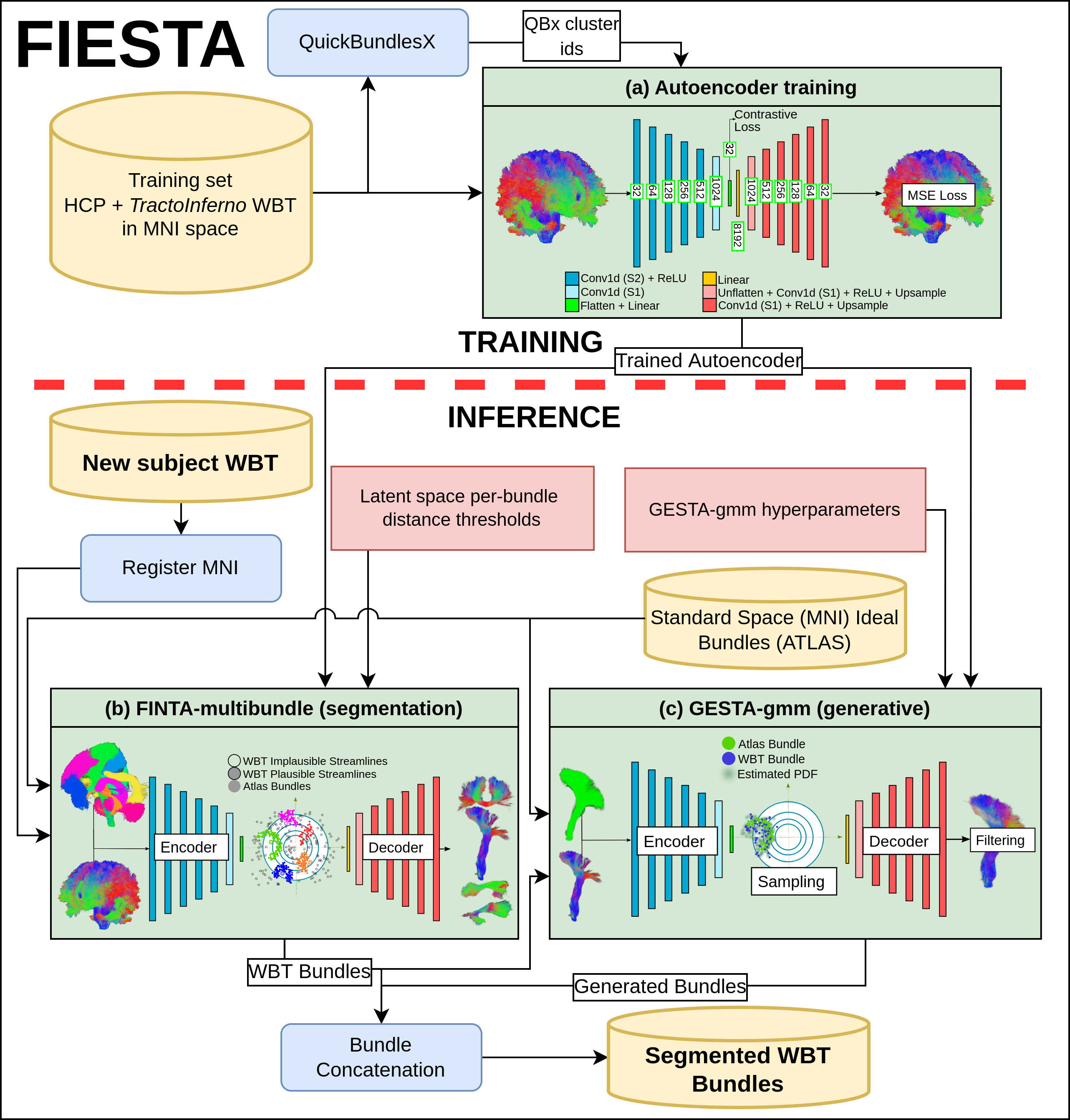}
    \caption{FIESTA pipeline. (a) Training of a convolutional autoencoder on raw dMRI whole-brain tractograms (WBT) with a combination of a mean squared error (MSE) loss and a contrastive loss. A streamline is a 1D signal with 3 channels. The contrastive loss' positive and negative pairs are based on \textit{QuickBundlesX} streamline cluster memberships\textcolor{black}{.} (b) The trained autoencoder is used to filter and bundle streamlines based on a k-nearest neighbors algorithm and a Euclidean distance threshold. The key idea behind the autoencoder's latent space is that plausible streamlines fall closer to the cleaned, filtered, and averaged bundles in the standard atlas than implausible streamlines\textcolor{black}{.} (c) A probability distribution function (PDF) is empirically estimated with a Parzen estimator \cite{bishop_pattern_2006} based on the embedding of each bundle and its atlas counterpart. Rejection sampling is used to generate new streamline samples from the estimated PDF. Sampled streamlines are then filtered by checking their fit to the underlying diffusion signal, a length range, a maximum winding angle, and a WM coverage rate, finally yielding a bundle with better spatial coverage.}
    \label{fig:ae_binta}
\end{figure}

FIESTA's three main underlying key concepts are summarized in Fig. \ref{fig:ae_binta}. First, an autoencoder is used to learn a streamline representation latent space (c.f. Fig. \ref{fig:ae_binta}a). Next, given a new tractogram and a reference set of bundles (typically provided by a trained specialist), \textcolor{black}{FINTA-multibundle} extracts false-positive-free bundles from a given whole-brain tractogram (c.f. Fig. \ref{fig:ae_binta}b). \textcolor{black}{ The tractogram is dissected by applying a filtering procedure to every bundle of interest individually (i.e., against all other groups, including implausible, and plausible streamlines belonging to other bundles).} Finally, \textcolor{black}{GESTA-gmm} populates each bundle by generating plausible streamlines supported by a filtering process to increase the spatial coverage (Fig. \ref{fig:ae_binta}c). FIESTA is a software suite built upon previous works that yields WM bundles from the concatenation of the output bundles of \textcolor{black}{FINTA-multibundle} and \textcolor{black}{GESTA-gmm}.

\subsection{Data}
\label{sec:data}
\textcolor{black}{Five} datasets were used for the development and the testing of FIESTA -- namely \textit{TractoInferno} \cite{poulin_tractoinferno_2022}, and the Human Connectome Project (HCP) \cite{glasser_minimal_2013, glasser_multi-modal_2016} as mentioned in Fig. \ref{fig:ae_binta}, and \textit{MyeloInferno} \cite{edde_high-frequency_2023}, \textcolor{black}{the Alzheimer's Disease Neuroimaging Initiative (ADNI) (\url{https://adni.loni.usc.edu}), and the Parkinson's Progression Markers Initiative (PPMI) (\url{www.ppmi-info.org/ access-data-specimens/download-data})} for the evaluation. The choice of datasets was based on several criteria, such as ease of access, multiple time-points per subject (\textit{MyeloInferno}\textcolor{black}{, ADNI, and PPMI}), high-quality data (HCP), and the availability of multiple WM tracking types (\textit{TractoInferno}). \textit{TractoInferno} and HCP data were used for the training of the convolutional autoencoder and early evaluation stage, while \textit{MyeloInferno} \textcolor{black}{, ADNI, and PPMI} were employed to evaluate the reliability of the pipeline. Also, \textit{TractoInferno} `\textbf{silver standard}' \cite{theaud_doris_2022} bundles were used for the threshold calibration steps. HCP subjects were also used for adjusting \textcolor{black}{GESTA-gmm}'s parameters.

\textit{TractoInferno} \cite{poulin_tractoinferno_2021} is a publicly available, widely heterogeneous database designed for machine learning purposes in dMRI tractography. It is composed of 354 subjects from 6 different datasets acquired with 5 different MRI scanners with various image resolutions, acquisition parameters, and subject ages. MRI acquisitions underwent a manual quality control (QC) process before using an ensemble of four tracking types -- namely local deterministic, local probabilistic, PFT~\cite{girard_towards_2014}, and SET~\cite{st-onge_surface-enhanced_2018} -- before being processed into \textit{RecoBundlesX}, yielding its silver standard. \textcolor{black}{\textit{TractoInferno} whole-brain tractograms were used for the autoencoder training (c.f. section \ref{sec:autoencoder_training}), whilst \textit{TractoInferno} silver standard bundles were only used for the latent space distance threshold calibration (c.f. section \ref{sec:threshold_calibration}.) } 

\textit{MyeloInferno} \cite{edde_high-frequency_2023} is a non-public dMRI dataset. \textcolor{black}{We have divided the \textit{MyeloInferno} dataset into 2 single time-point subsets and 2 five-time-point subsets. One single time-point subset, \textit{MyeloInferno-HC}, was} composed of \textcolor{black}{24} young and healthy \textcolor{black}{control (HC)} subjects (mean age 36 years $\pm$ 4.7 [standard deviation]; 17 women) \textcolor{black}{and the other, \textit{MyeloInferno-MS}, was composed of 21 young subjects with multiple sclerosis (MS) (mean age 38 years $\pm$ 6.8 [standard deviation]; 16 women)}. Written informed consent was obtained from participants and were recruited following the ethics protocol of the Centre de Recherche du Centre Hospitalier Universitaire de Sherbrooke (Sherbrooke, Canada). \textcolor{black}{Most HC and MS subjects in \textit{MyeloInferno} had 5 time-points or more. Thus, for the creation of our 2 test-retest subsets, we only kept data with 5 time-points or more and excluded time-points above 5 for our analysis. We named those 2 new subsets \textit{MyeloInferno-HC-TR} and \textit{MyeloInferno-MS-TR}. \textit{MyeloInferno-HC-TR} contains $N=18\times5=90$ acquisitions, whilst \textit{MyeloInferno-MS-TR} contains $N=19\times5=95$ acquisitions}. Thus, the data are ideal to evaluate the reproducibility of our pipeline. All images were acquired on a 3T Ingenia MR scanner with a 32-channel head coil (Philips Healthcare, Best, Netherlands) with T1w images acquired with a resolution of 1 mm isotropic voxels and dMRI images acquired with a resolution of 2 mm isotropic voxels, 100 gradient directions uniformly distributed over three shells ($b=\{300 (8), 2000 (32), 3000 (60)\} \mathrm{s/mm^2}$) (the number of directions per shell is in parentheses), and 7 unweighted images. Each tractogram was an ensemble of tractograms generated using the \textit{TractoFlow} pipeline \cite{theaud_tractoflow_2020}, which provided a PFT probabilistic tracking and a local probabilistic tracking that were concatenated. PFT tractograms were generated using interface seeding with 30 seeds per voxel, while local tractograms were generated using white matter seeding with 10 seeds per voxel. Both tracking methods were constrained with a streamline length range between 20 and 200 mm. \textcolor{black}{More details can be found in \citet{edde_high-frequency_2023} and on the corresponding website at \url{https://high-frequency-mri-database-supplementary.readthedocs.io/en/latest/index.html}.}

The Human Connectome Project (HCP) Young Adult dataset \cite{glasser_minimal_2013, glasser_multi-modal_2016} is a dataset of approximately 1200 subjects (age range 22-35 y/o) composed of different MRI modalities \textcolor{black}{acquired on Siemens scanners}. For the purposes of this project, only the T1w and dMRI acquisitions were used. All images were acquired on a 3T MR scanner with T1w images acquired with a resolution of 0.7 mm isotropic voxels and dMRI images acquired with a resolution of 1.25 mm isotropic voxels, 270 gradient directions equally distributed over three shells ($b=\{1000, 2000, 3000\} \mathrm{s/mm^2}$) and 6 unweighted images. Again, the tracking was done using the \textit{TractoFlow} pipeline \cite{theaud_tractoflow_2020} based on fODF estimated using $b=\{0, 1000, 2000, 3000\} \mathrm{s/mm^2}$ shells, which provided a PFT probabilistic tracking and a local probabilistic tracking that were concatenated. PFT tractograms were generated using interface seeding with 60 seeds per voxel, while local tractograms were generated using white matter seeding with 30 seeds per voxel. Both tracking methods were constrained with a streamline length range between 20 and 200 mm.

\textcolor{black}{The Alzheimer's Disease Neuroimaging Initiative (ADNI) dataset (\url{https://adni.loni.usc.edu}) was used in the preparation of this article. The ADNI was launched in 2003 as a public-private partnership, led by Principal Investigator Michael W. Weiner, MD. The primary goal of ADNI has been to test whether serial magnetic resonance imaging (MRI), positron emission tomography (PET), other biological markers, and clinical and neuropsychological assessment can be combined to measure the progression of mild cognitive impairment (MCI) and early Alzheimer's disease (AD). For this study, T1w and dMRI acquisitions were used. All images were acquired on a 3T MR scanner from different vendors with T1w images acquired with a resolution of 1 mm isotropic voxels and dMRI images acquired with a resolution of 2 mm isotropic voxels, 114 gradient directions distributed over three shells ($b=\{500 (6), 1000 (48), 2000 (60)\} \mathrm{s/mm^2}$) (the number of directions per shell is in parentheses) and 12 unweighted images. Tracking was done using the \textit{TractoFlow} pipeline \cite{theaud_tractoflow_2020} based on fODF estimated using $b=\{0, 1000, 2000\} \mathrm{s/mm^2}$ shells, which provided a PFT probabilistic tracking and a local probabilistic tracking that were concatenated. PFT tractograms were generated using interface seeding with 4 seeds per voxel, while local tractograms were generated using white matter seeding with 3 seeds per voxel. Both tracking methods were constrained with a streamline length range between 20 and 200 mm. For our analysis we created a test-retest subset (ADNI-TR) comprised of 23 subjects with 5 time-points each (baseline, 3 months, 6 months, 12 months, 24 months) yielding a total of 115 images. Also, we created a single time-point subset (ADNI-HC) with 21 old healthy subjects.}

\textcolor{black}{The Parkinson's Progression Markers Initiative (PPMI) database (\url{www.ppmi-info.org/ access-data-specimens/download-data}) was also used for this article. For this study, T1w and dMRI acquisitions were used. All images were acquired on a 3T MR scanner with T1w images acquired with a resolution of 1 mm isotropic voxels and dMRI images acquired with a resolution of 2 mm isotropic voxels, 64 gradient directions distributed over 1 shells ($b=1000 \mathrm{s/mm^2}$) and 1 unweighted images. Again, the tracking was done using the \textit{TractoFlow} pipeline \cite{theaud_tractoflow_2020} based on fODF estimated using $b=\{0, 1000\} \mathrm{s/mm^2}$ shells, which provided a PFT probabilistic tracking and a local probabilistic tracking that were concatenated. PFT tractograms were generated using interface seeding with 4 seeds per voxel, while local tractograms were generated using white matter seeding with 3 seeds per voxel. Both tracking methods were constrained with a streamline length range between 20 and 200 mm. 34 subjects with 3 time-points each were used (baseline, 12 months, 24 months) yielding a total of 102 images used from this dataset.}

\textit{TractoFlow} \cite{theaud_tractoflow_2020} was used to process HCP, \textit{MyeloInferno}\textcolor{black}{, ADNI, and PPMI} subjects, yielding PFT and local probabilistic whole-brain tractograms, while \textit{TractoInferno} authors were reached to get access to original raw tractograms (PFT, SET \cite{st-onge_surface-enhanced_2018}, local deterministic tracking and local probabilistic tracking).

\subsubsection{Bundle Atlas}
\label{sec:rbx}

\begin{figure}[!ht]
    \includegraphics[width=1\textwidth]{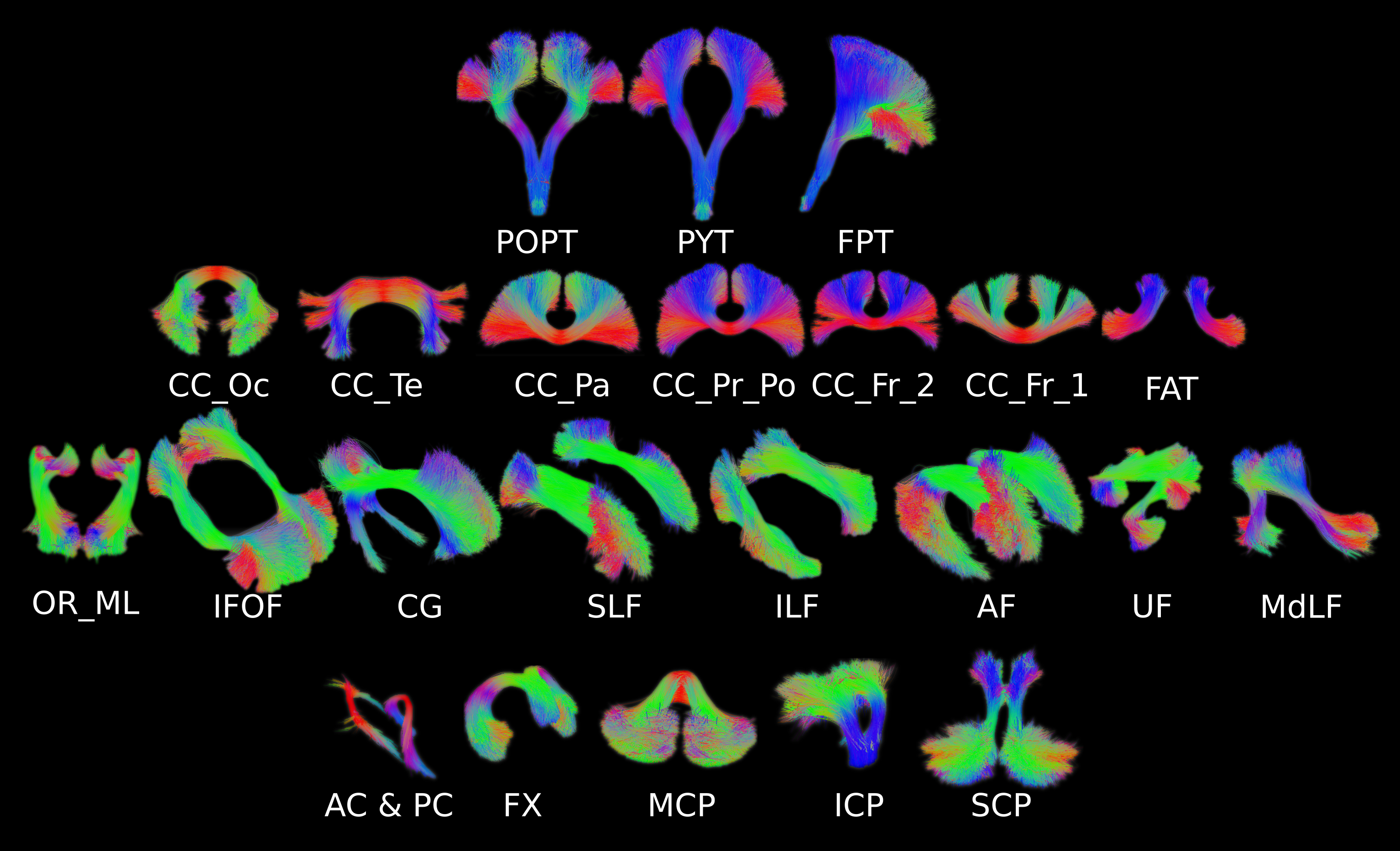}
    \caption{PAWM atlas bundles used for FIESTA development and evaluation. All symmetric bundles are joined together (left and right). Abbreviations are described in full detail in section \ref{sec:bundles}.}
    \label{fig:atlas}
\end{figure}

As shown in Fig. \ref{fig:ae_binta}, standard space ideal bundles (i.e., an atlas) are needed as a reference for FIESTA to work properly. For the development and evaluation of the current pipeline, we used an in-house but yet public Population Average of WM (PAWM) atlas to evaluate our framework \cite{rheault_population_2021}. The PAWM atlas was built in the context of \textit{RecoBundlesX} works by \citet{garyfallidis_recognition_2018, rheault_analyse_2020} based on ExTractor \cite{petit_structural_2022} and well-curated by a neuroanatomist to finally have bundles that fit their normative shapes \cite{rheault_population_2021}. Fig. \ref{fig:atlas} presents a representation of all the bundles in the PAWM atlas. Full names, abbreviations, and labels of bundles used in this work are indicated in Supplemental section \ref{sec:bundles}, whilst more context on the method used to generate these bundles is given in Supplemental section \ref{sec:pawm_atlas_supp}.

\subsection{Tractogram embedding}
\label{sec:tractogram_embedding}
Fig. \ref{fig:ae_binta}a shows a summary of the autoencoder's embedding and training scheme of the whole human brain streamlines. Autoencoders~\cite{hinton_reducing_2006} are a type of neural network designed to compress and reconstruct input data as faithfully as possible. They are made of an encoder (in blue) and a decoder (in red), two 1D convolutional neural networks in our case. The encoder projects a sample from the input space to a latent space, and the decoder reverses that operation. An undercomplete autoencoder~\cite{goodfellow_deep_2016} is one whose latent representation dimensionality is smaller than that of the input data. This forces the autoencoder to learn a latent space which embeds the most salient features of the input data. A well-constructed autoencoder encodes streamlines with similar properties (i.e., shapes, anatomical location, etc.) close to each other in the latent space.

Let $\mathcal{S}=\{ S_1, ..., S_N\}$ be a set of $N$ streamlines \textcolor{black}{non-linearly} registered in a common space (i.e., MNI \cite{fonov_unbiased_2009, fonov_unbiased_2011} \textcolor{black}{using T1w image-based registration} for the current pipeline) with $S_i=[\textbf{s}_i^1, ..., \textbf{s}_i^k]$ where $k$ is the number of vertices per streamline and $\textbf{s}_i^j \in \mathbb{R}^3$. By making sure that $k = D$ for all streamlines, with $D$ being a constant, we train a 1D convolutional autoencoder on raw whole-brain tractograms to learn a useful WM streamline latent representation. The training loss contains two terms: an MSE reconstruction loss and a contrastive loss to help the \textcolor{black}{bundle segmentation} problem.


The contrastive loss runs over pairs of samples $S_i, S_j$ to enforce streamlines of the same bundle to be as close a possible in the latent space. Following \citet{hadsell_dimensionality_2006}, the contrastive loss is built upon a binary label $y$ where $y=0$ if $S_i$ and $S_j$ have similar properties and $y=1$, if $S_i$ and $S_j$ are dissimilar. Also, let the parameterized Euclidean distance between two streamlines be defined as
\begin{equation}
D_\theta (S_i, S_j)=\| p_\theta (S_i) - p_\theta (S_j) \|_2,
\end{equation}

where $p_\theta$ is the encoder function. The contrastive loss function is 
\begin{equation}
    \mathcal{C}(\theta) = \sum_{i=1}^N C(\theta, y, S_i, S_j),
\end{equation}
with
\begin{equation}
C(\theta, y, S_i, S_j) = (1-y)C_S(D_\theta) + yC_D(D_\theta).
\end{equation}

$C_S$ and $C_D$ are designed such that minimizing $C$ w.r.t $\theta$ would result in low values of $D_\theta$ when $S_i, S_j$ are within the same bundle and high values for $D_\theta$ otherwise. Therefore,
\begin{equation}
C(\theta, y, S_i, S_j) = (1-y)\frac{1}{2}(D_\theta)^2 + y\frac{1}{2}\{max(0, m - D_\theta)\}^2,
\end{equation}

where the contrastive loss margin was set to $m=1.25$, as in \citet{hadsell_dimensionality_2006}.

Since whole-brain tractograms come with no annotation, similar and dissimilar pairs of streamlines are determined based on \textit{QuickBundlesX} clusters \cite{garyfallidis_quickbundlesx_2015} obtained with \{40, 30, 20, 10\} mm as input parameters. Thus, similar (or positive) streamline pairs are taken from the same \textit{QuickBundlesX} cluster, whilst dissimilar (or negative) streamline pairs are taken from two different clusters.

Finally, the MSE reconstruction loss is defined as 
\begin{equation}
\mathcal{M}(\theta, \phi) = \sum_{i=1}^N \| \hat{S}_i - S_i \|_2^2,
\end{equation}
%
%
%
%
where $\hat{S}_i = q_\phi(p_\theta(S_i))$ is a reconstructed streamline and $q_\phi(\cdot)$ is the decoder. Therefore, the overall training loss is 
\begin{equation}
\mathcal{L}(\theta, \phi) = \mathcal{M}(\theta, \phi) + \lambda \mathcal{C}(\theta),
\end{equation}
with $\lambda$ being a hyperparameter. The hyperparameter $\lambda$ was empirically determined to approximately balance the MSE loss with the contrastive loss on a randomly initialized autoencoder. In our case, the optimal value was $\lambda=400$. 


\subsubsection{The autoencoder training scheme}
\label{sec:autoencoder_training}
We trained the autoencoder in a train-validation-test scheme. Our whole dataset is composed of 120 subjects randomly shuffled with an 80/20 split between \textit{TractoInferno} and HCP datasets (c.f. section \ref{sec:data}). The train/validation/test split was therefore 100/10/10. \num[group-separator={,}]{200000} streamlines per subject were used, with \num[group-separator={,}]{100000} streamlines per HCP whole-brain tractogram ($N=2$ -- namely PFT and local probabilistic tracking) and \num[group-separator={,}]{50000} streamlines per \textit{TractoInferno} whole-brain tractogram were randomly sampled from each raw tractograms ($N=4$ -- namely PFT, SET, local deterministic tracking and local probabilistic tracking). In total, our dataset contained \num[group-separator={,}]{24000000} streamlines. Also, each streamline is resampled to a fixed dimensionality of $D=256$ 3D vertices and the latent space dimensionality is fixed to $d=32$.

\subsection{Tractogram Segmentation}
\label{sec:tractogram_filtering_and_clustering}
Once the autoencoder is trained, one can use it to filter and bundle a new tractogram. As displayed in Fig. \ref{fig:ae_binta}b, every streamline that ought to be filtered and segmented is \textcolor{black}{non-linearly} registered into the MNI reference space \textcolor{black}{using T1w image-based registration} and projected into the latent space using the encoder neural network. This is done alongside the bundle streamlines of the atlas. 
A k-nearest neighbors (k-NN) algorithm is used to assign each streamline (in the latent space) to the majority class label among the nearest atlas streamline (also in the latent space). The absolute Euclidean distance between the WM streamline, and its atlas counterpart is then compared to a pre-determined threshold. To determine the threshold that best balances true positives and false positives, a ROC curve was used, as suggested by \citet{legarreta_filtering_2021}, alongside manual adjustment (c.f. section \ref{sec:threshold_calibration}). Manual adjustments is mandatory to qualitatively match atlas bundles. Thus, if the distance is smaller than the threshold, the streamline is kept for downstream tasks. Otherwise, the streamline is considered implausible and, therefore, discarded.

\subsubsection{Latent space distance threshold calibration steps}
\label{sec:threshold_calibration}

\begin{figure}[!ht]
    \centering
    \includegraphics[width=0.75\textwidth]{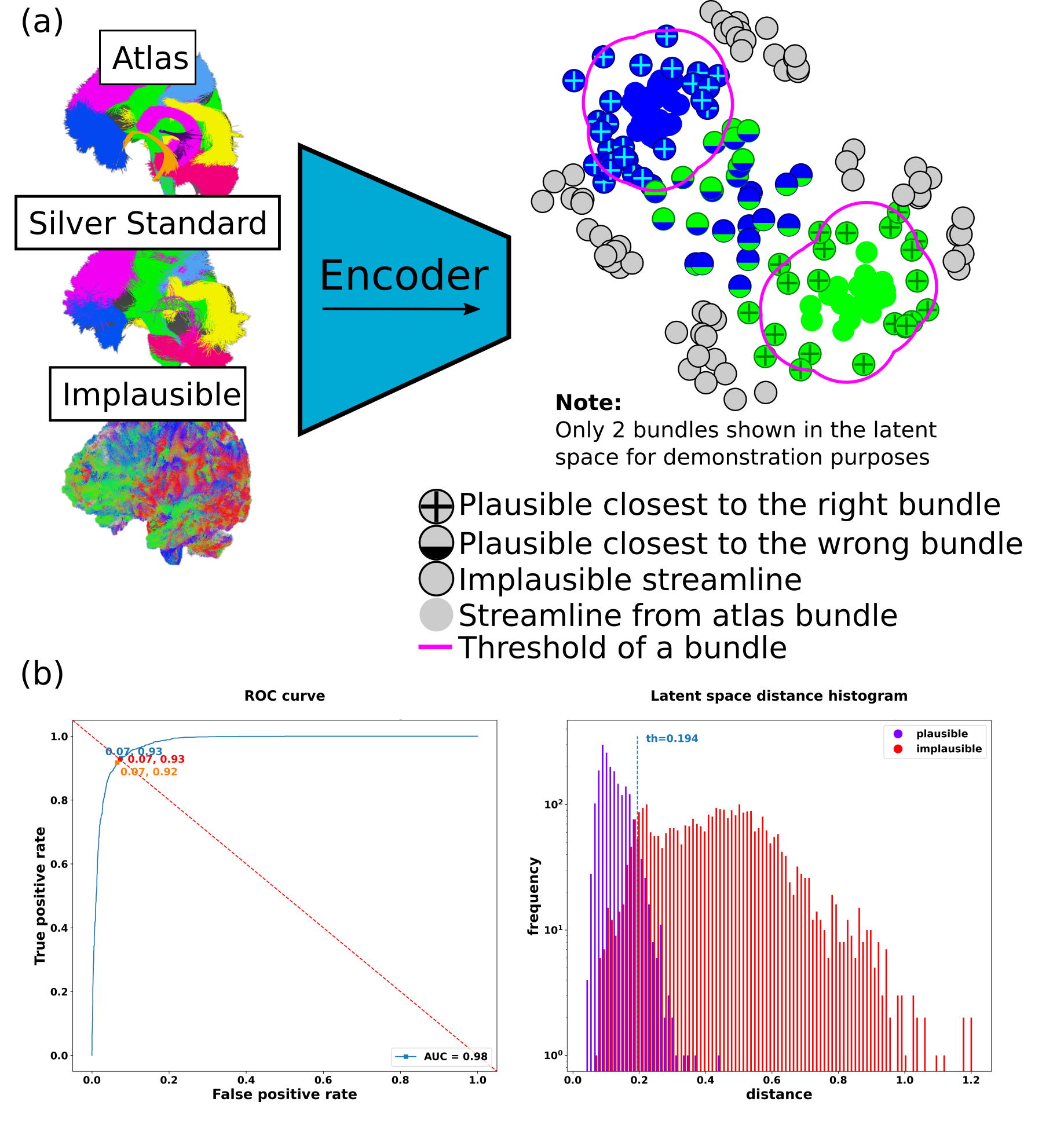}
    \caption{Threshold calibration steps where (a) atlas, silver standard, and implausible streamlines in MNI space are projected in the latent space. \textit{\textcolor{black}{Near-}optimal} thresholds are found by assigning all silver standard and implausible streamlines to the class of the closest atlas streamline and by balancing true positives and false positives for each bundle. (b) For each bundle in the atlas, the inverse diagonal of the ROC curve, built from the right silver standard streamlines assigned to the current atlas bundle (interpreted as positive streamlines) and the wrong silver standard combined with the implausible streamlines assigned to the current atlas bundle (interpreted as negative streamlines), is used to find the \textit{\textcolor{black}{near-}optimal} threshold. \textbf{Left:} \textit{\textcolor{black}{near-}optimal} threshold for 1 bundle found at the intersection of the ROC curve with the inverse diagonal and \textbf{Right:} histogram of one bundle with the x-axis giving the distance to the closest streamline in the atlas to each streamline in the silver standard (plausible) and implausible streamlines. The \textit{\textcolor{black}{near-}optimal} distance threshold found with the ROC curve analysis is indicated.}
    \label{fig:threshold_optimal}
\end{figure}

\begin{figure}[!ht]
    \centering
    \includegraphics[width=1\textwidth]{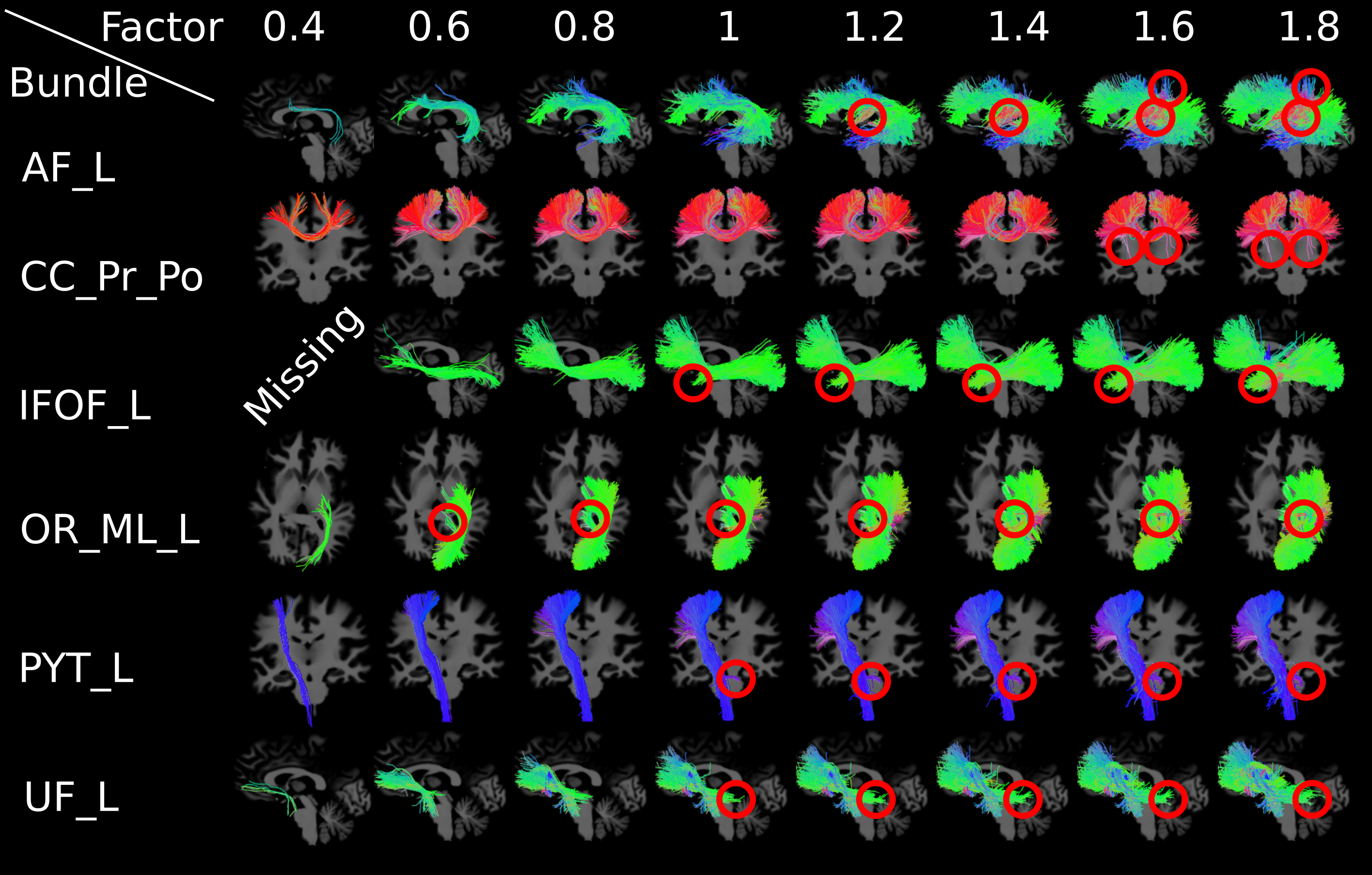}
    \caption{Effect of various factors based on automated thresholds found with ROC curves analysis where a factor of 1 represents the original thresholds. Factors below 1 mean that original thresholds were reduced, while factors above 1 mean that original thresholds were increased. Red circles indicate false positive streamlines according to atlas ideal bundles presented in Fig. \ref{fig:atlas}. \textit{Missing} means that the method did not extract any streamline for a particular bundle.}
    \label{fig:thresholds}
\end{figure}

\textcolor{black}{Bundle segmentation} methods require a calibration to account for each bundle's unique shape; for example, with \textit{RecoBundles} and \textit{RecoBundlesX}, there are multiple parameters to set for each bundle. Fortunately, \textcolor{black}{FIESTA has only} one threshold per bundle (c.f. Fig. \ref{fig:thresholds} to see threshold effect on bundles). For calibration, a \textit{near-optimal}  per-bundle threshold is determined automatically followed by manual adjustments to obtain the desired bundle shape (c.f. Fig. \ref{fig:threshold_optimal}). \textit{Near-optimal} thresholds were found by equally distributing the per-bundle true positive and false positive streamlines, using the ROC curve analysis method presented by \citet{legarreta_filtering_2021}. We used \textit{TractoInferno} silver standard \cite{theaud_doris_2022} bundles in the validation set as true positive streamlines. We consider as implausible the streamlines from the raw tractogram that are absent from the silver standard bundles. \textcolor{black}{To avoid any confusion for the latent space calibration step, the following list explains each data used.} 

\textcolor{black}{\begin{enumerate}
    \item \textbf{IST}: Implausible streamlines from the TractoInferno validation set;
    \item \textbf{SST}: Silver standard streamlines from the TractoInferno validation set;
    \item \textbf{WBTT}: Whole-brain tractograms from the TractoInferno validation set, where $IST + SST = WBTT$;
    \item \textbf{PAWM}: Ideal atlas of bundles comprised of plausible streamlines (c.f. section \ref{sec:rbx}).
\end{enumerate}}

The bundle-wise thresholds are thus determined as follows:

\begin{enumerate}
    \item Encode atlas bundles \textcolor{black}{(PAWM)} and their flipped versions \textcolor{black}{(to make the model independent of its direction)} into latent vectors; 
    \item Encode plausible silver standard bundles \textcolor{black}{(SST)} and implausible streamlines \textcolor{black}{(IST)} and their flipped version from the validation set;
    \item For each latent vector from the validation set (both plausible \textcolor{black}{(SST)} and implausible \textcolor{black}{(IST)}), find the closest streamline in the atlas using a k-NN algorithm. This step forms groups of \textcolor{black}{candidate} plausible and implausible streamlines with a class label;
    \item For each group of streamlines, optimize the Euclidean distance threshold that maximizes the number of true positive streamlines and minimize the number of false positive streamlines. Note that during this stage, \textcolor{black}{candidate} plausible streamlines assigned to the wrong class are considered implausible. This step is equivalent to taking the threshold value at the intersection between each ROC (bundle-wise) curve and the inverse diagonal line;
    \item Qualitatively manually adjust the \textit{\textcolor{black}{near-}optimal} thresholds, if needed, to improve atlas bundle similarity with the validation set bundles.
\end{enumerate}

To understand the effect of the choice of threshold, we visually inspected the shape and the quality of the resulting bundles. We set up an experiment with a subset of the bundles from the atlas presented in section \ref{sec:rbx}. We studied the effect of 8 uniformly spaced threshold values centered on the \textit{\textcolor{black}{near-}optimal} threshold of the \texttt{AF\_L}, the \texttt{CC\_Pr\_Po}, the \texttt{IFOF\_L}, the \texttt{OR\_ML\_L}, the \texttt{PYT\_L} and the \texttt{UF\_L} bundles. Fig. \ref{fig:thresholds} illustrates the results of that experiment. We set the values using 20\% increments based on the ROC curves analysis thresholds. Therefore, it is possible to observe that low thresholds seems to increase bundle specificity at the expense of the sensitivity. On the other hand, high threshold values, classifying more streamlines as positives, increase the sensitivity at the expense of the bundle specificity. Red circles in Fig. \ref{fig:thresholds} indicate streamlines that should not be part of final bundles based on the atlas ideal bundles. Interestingly, such streamlines are not always present when the threshold is higher than the \textit{\textcolor{black}{near-}optimal} threshold. In fact, we see that spurious streamlines are still present in the \texttt{OR\_ML} even with a 40\% threshold reduction. Thus, this analysis shows that threshold manual adjustment is mandatory, and threshold values based only on ROC curves analysis should not all be used as is. In our case, for some bundles, we needed to decrease the latent space distance threshold, thus increasing the specificity, to more closely match, qualitatively, bundles from the PAWM atlas (c.f. Fig \ref{fig:atlas}).

\subsection{Bundle coverage improvement using generative sampling (\textcolor{black}{GESTA-gmm})}
\label{sec:bundle_coverage_improvement}
As shown in Fig. \ref{fig:ae_binta}c, \textcolor{black}{GESTA-gmm} is used, after \textcolor{black}{FINTA-multibundle}, to better populate each bundle. This module compensates the imperfections of the latent space-based threshold segmentation method and the missing streamlines in the original whole-brain tractogram, especially in hard-to-track bundles~\cite{rheault_bundle-specific_2019}. Thus, for each bundle in the atlas, we \textcolor{black}{consider} the streamlines \textcolor{black}{labeled} by \textcolor{black}{FINTA-multibundle} \textcolor{black}{as belonging to} the current bundle (see Fig. \ref{fig:ae_binta}b). An input ratio $\textcolor{black}{1}{:}\textcolor{black}{1}$ between the number of streamlines to use from each bundle (atlas and \textcolor{black}{FINTA-multibundle} output, c.f. Supplemental section \ref{sec:gesta_ratio} for more details) is specified alongside the maximum total number of streamlines to use. The decision to estimate the latent space PDF with streamlines from the subject bundles and the atlas bundles was motivated by the fact that ideal bundles are built to fully cover the target anatomical region. Therefore, missing streamlines from the original bundles can be recovered if a portion of the streamline seeds \cite{legarreta_generative_2022} for the latent space PDF estimation is based on the atlas bundles. Then, all streamlines used to estimate the PDF are embedded with the encoder. The PDF is empirically estimated using a Parzen estimator \cite{bishop_pattern_2006} over all embedded points with a Gaussian kernel. The kernel bandwidth is automatically estimated using Silverman's rule of thumb~\cite{silverman_density_1986}.


As described in \cite{legarreta_generative_2022, painchaud_cardiac_2020}, since streamline latent representations live in a locally linear manifold, rejection sampling (RS) \cite{bishop_pattern_2006} can be used to sample new data from an empirically estimated PDF. Thus, giving a set of latent vectors $\mathcal{Z}$ with $\textbf{z}_i \in \mathbb{R}^d, \textbf{z}_i \subset \mathcal{Z}$ and $\textbf{z}_i = p_\theta(S_i)$, our goal is to estimate a new set of latent vectors $\mathcal{Z}' \not\subset \mathcal{Z}$ where the new PDF of estimated vectors $P(\textbf{z}')$ is close to $P(\textbf{z})$. As described previously, the PDF of $P(\textbf{z})$ is unknown, and we estimate it with a Parzen estimator. Because $P(\textbf{z})$ is difficult to sample, we sample an easier PDF $Q(\textbf{z})$. In our case, $Q(\textbf{z})$ is a mixture of Gaussians estimated with the expectation-maximization algorithm~\cite{bishop_pattern_2006}. 
The RS procedure works as follows: one first generate a random sample from $Q(\textbf{z})$ as well as a random number $u_0$ i.i.d. of a uniform distribution between $[0, KQ(\textbf{z})]$ where $K$ is a constant.  If $u_0 > P(\textbf{z})$, the sample is rejected, otherwise it is accepted. 

Finally, the accepted vectors are decoded to generate new streamlines. 
As described in GESTA~\cite{legarreta_generative_2022}, those streamlines need to fit certain anatomical constraints. Therefore, we adopt the four proposed constraints – namely a length range, a WM coverage, a maximum curving angle, and the local streamline orientation to fODF peak angle (c.f. Supplemental section \ref{sec:dmri_filters}). Furthermore, we trim off the vertices at each end of generated streamlines that overshoots the gray matter with a WM mask. The final bundles used for evaluation are the \textcolor{black}{simple} concatenation \textcolor{black}{(without further filtering)} of \textcolor{black}{FINTA-multibundle} and \textcolor{black}{GESTA-gmm} bundles.

\subsubsection{Number of generated streamlines}
\label{sec:number_of_generated_streamlines}

\begin{figure}[!ht]
    \centering
    \includegraphics[width=1\textwidth]{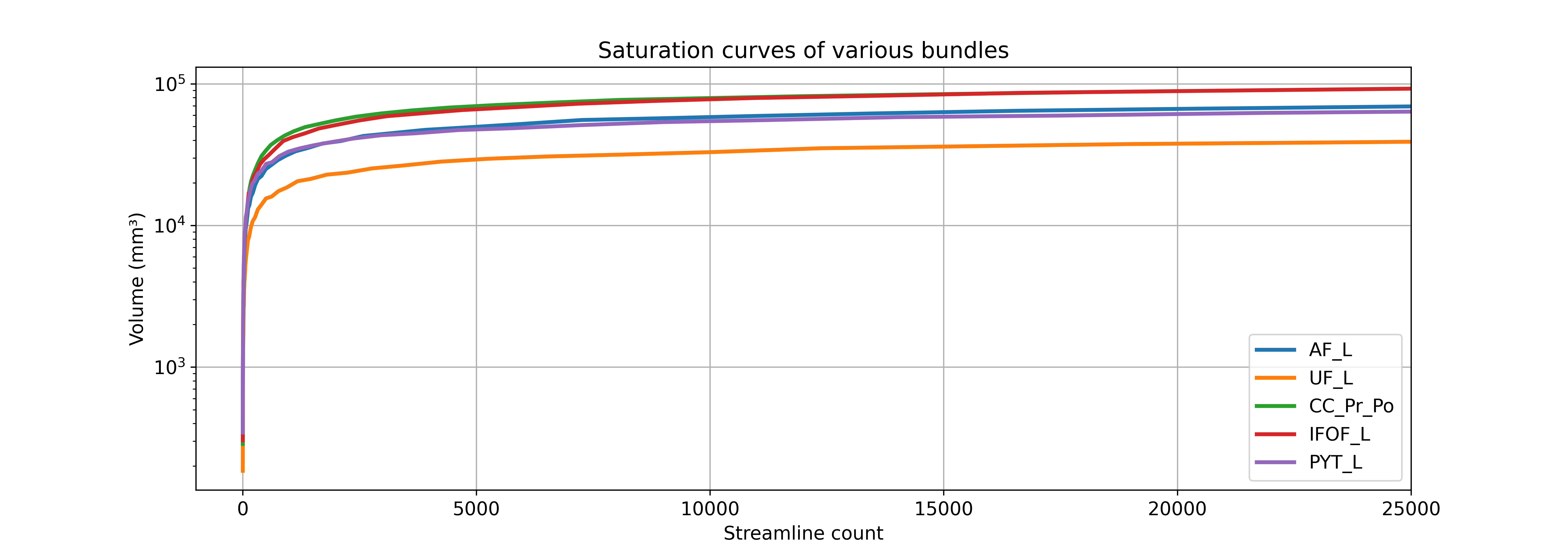}
    \caption{Saturation curves of various bundles generated by \textcolor{black}{GESTA-gmm} showing that between \num[group-separator={,}]{5000} and \num[group-separator={,}]{15000} is a reasonable amount of generated streamlines to cover the entirety of the bundle. After \num[group-separator={,}]{15000} streamlines, the volume of each bundle does not increase significantly.}
    \label{fig:saturation_curves}
\end{figure}

To assess the capacity of \textcolor{black}{GESTA-gmm} to generate diverse streamlines that ``fill'' the anatomy well, we produced saturation curves \cite{gauvin_assurance_2016, rheault_analyse_2020}, which plot the bundle volume with respect to the streamline count. We expect that if adding new streamlines does not affect the bundle volume significantly, the bundle is saturated, i.e., it is ``well filled''. We analyzed saturation curves \cite{gauvin_assurance_2016, rheault_analyse_2020} of bundle volumes from the \textcolor{black}{GESTA-gmm} process compared to the streamline count over 5 bundles based on a whole-brain tractogram of 1 million streamlines. To generate such saturation curves, we produced bundles with \num[group-separator={,}]{25000} streamlines \textbf{post \textcolor{black}{GESTA-gmm} filtering}. After, we logarithmically randomly sampled with replacement 50 subsampled bundles from each bundle. The volume was then computed for each sub-bundle. Fig. \ref{fig:saturation_curves} presents the saturation curves for those 5 bundles – namely the \texttt{AF\_L}, the \texttt{UF\_L}, the \texttt{CC\_Pr\_Po}, the \texttt{IFOF\_L}, and the \texttt{PYT\_L}. We see that, after~\num[group-separator={,}]{5000} generated streamlines, each bundle volume starts to saturate and after~\num[group-separator={,}]{15000}, no real volume gain is obtained from the generative process. Therefore, we fixed the number of latent-sampled vectors to \num[group-separator={,}]{25000} prior to decoding in our experiments, where about half the generated bundles had a final count of more than \num[group-separator={,}]{5000} streamlines. Finally, bundles with less final streamlines were generally caused, not by a wrongful generative process, but by imperfect WM masks or bundle atlas alignments with the underlying standard space anatomy.


\subsection{Reliability evaluation metrics}
\label{sec:metrics}
FIESTA's test-retest reliability was assessed using the \textit{MyeloInferno} dataset 
according to five overlap metrics – namely \begin{enumerate*}[label=(\roman*)] \item the voxel-wise Dice coefficient, \item the weighted voxel-wise Dice coefficient, \item the voxel-wise bundle adjacency, \item the streamline-wise bundle adjacency, and \item the streamline density correlation \end{enumerate*} \cite{rheault_tractostorm_2020, cousineau_test-retest_2017, dice_measures_1945}. The Dice and the weighted Dice coefficients are overlap measures between 2 bundles varying from 0 to 1 where 1 means a perfect overlap. The weighted Dice score is weighted by the number of streamlines in each voxel. Bundle adjacency measures are distance metrics, similar to the Hausdorff distance \cite{rockafellar_variational_2009}, which yield an average distance between 2 bundles. Metrics are reported in mm and, while there is no cap for those metrics, a value of 0 mm means that there is no distance between 2 bundles. Finally, the streamline density correlation yields a Pearson correlation between 2 bundle density maps going from 0 to 1 with 1 meaning that those 2 bundles are perfectly correlated. Intra-subject tractography variability was assessed pairwise, i.e., for a total of $J$ acquisitions per subject, metrics were computed between images 
$[1\rightarrow2, 1\rightarrow3, 1\rightarrow4, ..., J-1\rightarrow J]$ 
equaling to a total of $J(J-1)/2$ comparisons per bundle. Also, to gauge FIESTA's reproducibility 
we computed the intraclass correlation coefficient (ICC) over volume, and average bundle lengths. Based on Koo and Li (2016) \cite{koo_guideline_2016}, ICC estimates and their 95\% confidence intervals were computed based on a two-way mixed-effects, consistency, multiple raters/measurements model (ICC (3,k)). Implementation was done using the Pingouin statistical Python package (v0.5.1) \cite{vallat_pingouin_2018}.

\subsection{Hyperparameters}
The final fixed hyperparameters used for FIESTA were a generative sampling of \num[group-separator={,}]{25000} new streamlines based on \num[group-separator={,}]{10000} seed points from an atlas/subject bundle ratio of $\textcolor{black}{1}{:}\textcolor{black}{1}$ followed by the \textcolor{black}{GESTA-gmm} filtering process. It should also be noted that, for bundles where \textcolor{black}{FINTA-multibundle} was not able to give 50\% of the \num[group-separator={,}]{10000} required seeds, atlas bundle streamlines were used to compensate. 

\subsection{Methods' comparison}
To have a broad range of comparisons, we benchmarked our method against 5 state-of-the-art \textcolor{black}{bundle segmentation} methods –- namely \textit{RecoBundles} (RB) \cite{garyfallidis_recognition_2018}, \textit{RecoBundlesX} (RBx) \cite{rheault_analyse_2020}, \textit{TractSeg} \cite{wasserthal_tract_2018, wasserthal_tractseg_2018, wasserthal_combined_2019, wasserthal_multiparametric_2020}, \textit{XTRACT} \cite{warrington_xtract_2020}, and \textit{White Matter Analysis} (WMA) \cite{odonnell_automatic_2007, odonnell_unbiased_2012, zhang_anatomically_2018}. We also compared \textcolor{black}{FINTA-multibundle} against FIESTA (\textcolor{black}{FINTA-multibundle} + \textcolor{black}{GESTA-gmm}) to see the effect of generative sampling on bundles' reliability. While \textit{RecoBundles}, \textit{RecoBundlesX}, WMA, \textcolor{black}{FINTA-multibundle}, and FIESTA used the whole-brain tractogram from \textit{TractoFlow}, we used the default tracking algorithm for \textit{TractSeg} and \textit{XTRACT}. Also, as \textit{XTRACT} software yields bundle density maps, the evaluation was done on thresholded maps with a value of 0.5\% of the highest bundle value. WMA whole-brain tractograms had to be downsampled to \num[group-separator={,}]{500000} streamlines due to algorithmic computational limitations. In comparison, \textit{RecoBundles}, \textit{RecoBundlesX}, \textcolor{black}{FINTA-multibundle}, and FIESTA whole-brain tractograms were not downsampled and used the $\sim$4M streamlines yielded by \textit{TractoFlow}. \textcolor{black}{A list of the main parameters for each method is presented in Supplemental section \ref{sec:parameters} alongside their approximate computation time for a standard desktop computer.}

Since the 5 \textcolor{black}{bundle segmentation} methods against which FIESTA is compared do not have the same bundle names and/or definitions, as a matter of fairness, only the similar bundles from all methods were compared together. 
Therefore, 27 similar bundles are kept for the analysis since some bundles in \textit{TractSeg}, WMA, and \textit{XTRACT} had completely different definitions than that of FIESTA. All compared bundles from each method can be found in Supplemental section \ref{sec:bundle_compared}. It should be noted that WMA misses the \texttt{SCP} bundle, while \textit{XTRACT} only shared 17 similar bundles compared to the other methods. Thus, the global average results were computed on those 17 bundles, whilst individual results were computed on the 27 bundles shared amongst the method.

\textcolor{black}{\subsection{FIESTA qualitative evaluation}
\label{sec:qualitative_evaluation_description}
To gain a better intuition into the different tested methods, we conducted a qualitative analysis on a random subject from the \textit{MyeloInferno\textcolor{black}{-HC-TR}} subset. We qualitatively compared various bundles from the different benchmarked methods in order to understand the pitfalls and the quality of each one of them. As one subject is not sufficient to conclude on the bundle's anatomical accuracy, we conducted a more extensive qualitative assessment using the dMRIQCpy toolbox \cite{theaud_dmriqcpy_2022} on the 95 individual images of the \textit{MyeloInferno-MS-TR} dataset. The qualitative evaluation was conducted using a 3 score classification following the user guide of dMRIQCpy on the 27 similar bundles shared amongst methods. The 3 possible scores were either \textbf{pass}, \textbf{warning}, or \textbf{fail}. For each bundle, or a pair of bundles if applicable, a \textbf{pass} score was given when the bundle clearly respected the anatomical definition of the atlas, overlapped the correct brain region, and was located in the expected hemispheric and anatomical region. A \textbf{warning} score was given if the bundles respected every previous condition but were sparsely populated. Finally, a \textbf{fail} score was given if any of the previous three conditions were not respected. We decided to do this test on a cohort of subjects with a neurological disease as a good test to understand the behavior of FIESTA on disease subjects.}

\textcolor{black}{\subsection{Generated fibers faithfulness}
\label{sec:gesta-usability}
To investigate the impact of generative sampling on tractography and the faithfulness of generated streamlines, we conducted two experiments. We used the 24 acquisitions from the \textit{MyeloInferno-HC} subset and the 21 from the ADNI-HC subset. All selected subjects were healthy and had no diagnosed neurological disease. In the first experiment, we aimed to evaluate the effectiveness of the 4 anatomical constraints, especially the constraint on the diffusion signal. Since we chose an atlas of bundles made of streamlines from young adults, we assumed that the number of streamlines rejected based on the 4 anatomical constraints should be greater on older adults than on younger ones. Two reasons pushes us to this hypothesis. First, because the MNI template brain was constructed based on young adults (aged 18.5–43.5 years) \cite{fonov_unbiased_2011}, the registration needed to bring streamlines from the native space to the template space for FIESTA to work properly should be more accurate on younger brains as brain structures are more similar. Second, as the streamlines from the atlas used for our tests were constructed from young adults, their shape should be closer to streamlines from young adult brains. Therefore, when such data is used to seed part of the autoencoder latent space for the GESTA-gmm module, the  estimated probability distribution should be closer to that of younger subjects. Thus, while sampling into this distribution, a greater percentage of generated streamlines should be accepted after passing the 4 constraints on young subjects.}

\textcolor{black}{The goal of the second experiment is to assess the impact of \textcolor{black}{GESTA-gmm} on the streamline-wise bundle adjacency metric between the 2 healthy populations, as this metric can be interpreted as how far 2 populations of streamlines are from each other. The goal is to avoid having a substantial closer streamline-wise bundle adjacency between young and old bundles. Thus, for every subject in the \textit{MyeloInferno-HC} population, we computed the streamline-wise bundle adjacency with every bundle of every subject in the ADNI-HC 21 subjects population. We repeated the experiment once on the bundles segmented by the \textcolor{black}{FINTA-multibundle} module, and once on the final output of FIESTA comprising the bundles from the \textcolor{black}{FINTA-multibundle} module and the \textcolor{black}{GESTA-gmm} module. As the number of streamlines will increase using the FIESTA output compared to the FINTA-multibundle output, we should expect a slight improvement in the streamline-wise bundle-adjacency, but it must not be substantial.}

\textcolor{black}{\subsection{Performance on disease cohorts}
In order to grasp the performance of FIESTA on disease subjects, we extended the test-retest analysis to subjects with neurodegenerative diseases. Therefore, the ADNI-TR, PPMI, and \textit{MyeloInferno-MS-TR} datasets were used with their multiple time-points per subject. ADNI-TR subjects used for this analysis included a mixture of normal aging, MCI, and AD subjects. PPMI subjects were only composed of subjects with Parkinson's disease (PD). \textit{MyeloInferno-MS-TR} contained subjects with multiple sclerosis disease. All three datasets contain multiple time-points, with 5 for ADNI-TR and \textit{MyeloInferno-MS-TR}, and 3 for PPMI. Thus, we repeated the reliability evaluation on those 3 datasets following the same procedure presented in section \ref{sec:metrics}}

\section{Results}
\label{sec:results}

\subsection{Qualitative Analysis}
\label{qualitative_analysis}
\begin{figure}[tp]
    \centering
    \includegraphics[width=1\textwidth]{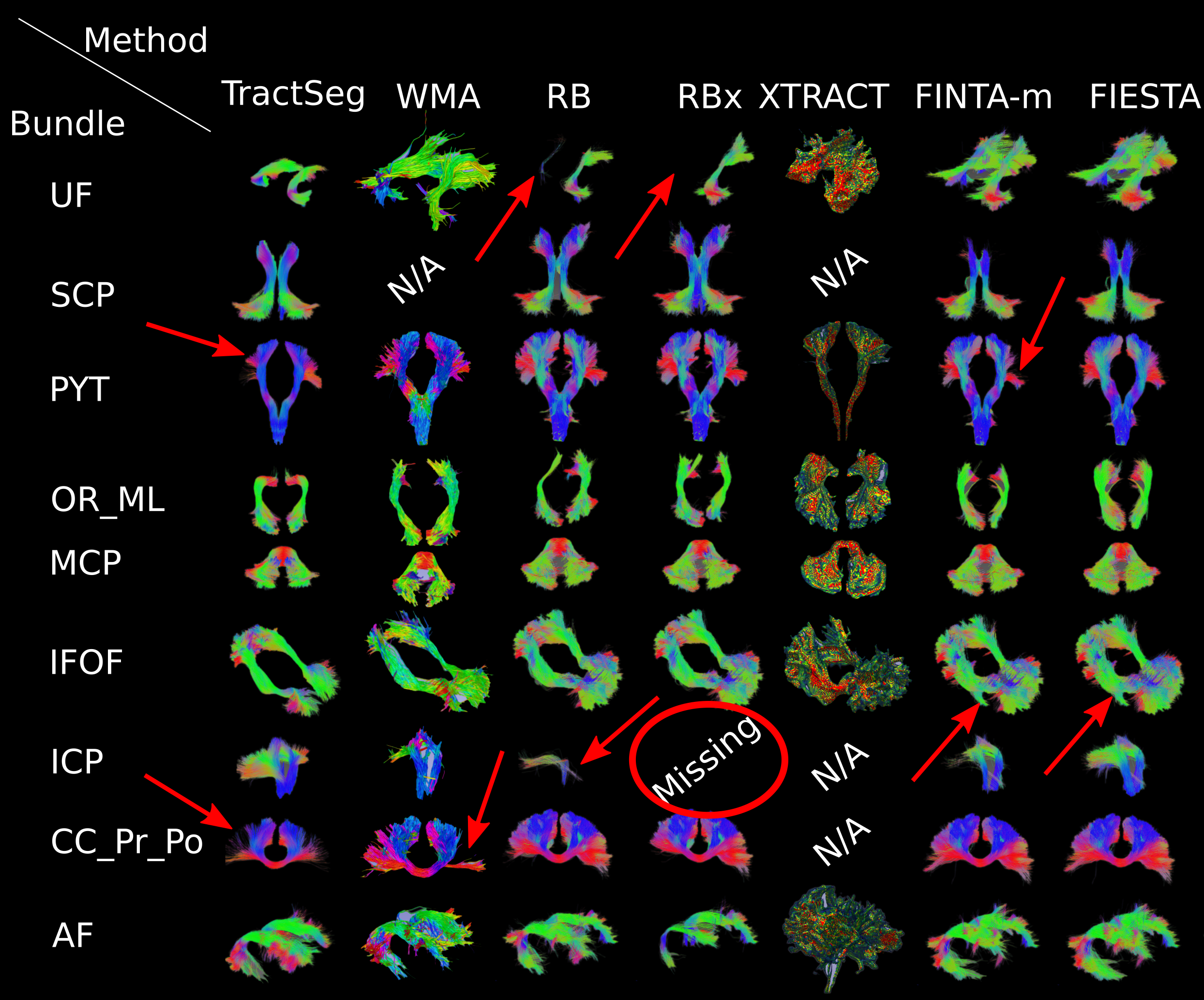}
    \caption{Qualitative results for a subset of bundles from the different benchmarked methods from one \textit{MyeloInferno} subject. 
    \textit{N/A} means that the bundle was not included in the method, while \textit{Missing} means that the method failed to extract that bundle. Lateral bundles (in both hemispheres) are shown jointly: \texttt{UF}, \texttt{SCP}, \texttt{PYT}, \texttt{OR\_ML}, \texttt{IFOF}, \texttt{ICP}, and \texttt{AF}. \textit{XTRACT} bundles show density maps (thresholded at 0.5\%). FINTA-m stands for \textcolor{black}{FINTA-multibundle. Red arrows point to suspected missing streamlines or wrongly bundled.}\vspace{-0.5cm}}
    \label{fig:methods}
\end{figure}

To gain insight into the different methods, we conducted a qualitative analysis on a random subject from the \textit{MyeloInferno\textcolor{black}{-HC-TR}} dataset. Fig. \ref{fig:methods} lays out the same bundles across each \textcolor{black}{bundle segmentation} method for a given subject. First, it is qualitatively seen that FIESTA's results (which contains generated streamlines), seems to have the best hemispheric homogeneity. This is particularly true for the \texttt{PYT} and the \texttt{CC\_Pr\_Po} bundles. \textit{TractSeg} misses some WM streamlines on the left \texttt{PYT} and the \texttt{CC\_Pr\_Po} bundles. Otherwise, \textit{TractSeg} and WMA seem to output cleaned bundles with few spurious streamlines. The major drawback of WMA comes from its inability to process the original tractograms due to memory issues ($\sim$4M streamlines in the \textit{MyeloInferno} dataset for each whole-brain tractogram). \textit{RecoBundles} and \textit{RecoBundlesX} seem to be the least effective methods. For \textit{RecoBundles}, the left \texttt{UF} is almost empty, while the left \texttt{ICP} is missing. The same goes for \textit{RecoBundlesX} with a completely missing \texttt{ICP} (left and right) and a missing left \texttt{UF}. Also, comparing their \texttt{SCP} to that from the atlas (c.f. Fig. \ref{fig:atlas}), it is seen that several of those streamlines belong to the medulla, which were not included in the atlas bundles. \textit{XTRACT} retrieves almost all streamlines for each bundle. Unfortunately, \textit{XTRACT} has a high sensitivity but a low specificity for some bundles such as the \texttt{AF}, the \texttt{UF}, and the \texttt{OR}. Finally, \textcolor{black}{FINTA-multibundle} clusters correctly all bundles, but with an inhomogeneous fanning for the \texttt{PYT}. \textcolor{black}{The only qualitative issue found with FIESTA and FINTA-multibundle is seen on the \texttt{IFOF}. Indeed, some outlier streamlines seem to appear on the inferior part of the bundle.}

\textcolor{black}{Concerning the qualitative analysis made on 95 acquisitions of the \textit{MyeloInferno-MS-TR} dataset using the dMRIQCpy \cite{theaud_dmriqcpy_2022} toolbox, all the 27 evaluated bundles over the whole dataset were given a score of \textbf{pass}. Such results confirm that, qualitatively, FIESTA performs well on a disease cohort where all bundles are fully populated, located in the right hemisphere and in the correct anatomical region. During this exercise, the shape of all bundles were compared to the shape of the bundle in the PAWM atlas. All bundles presented a qualitative similar shape to the atlas of bundles used for the development of FIESTA.}

\subsection{Reliability Analysis}
\label{sec:reliability}

Table \ref{tab:average_overlap_metrics} presents the results on the {\em MyeloInferno} dataset of the average similarity metrics for each method benchmarked over 17 similar bundles, all shared among the different methods.  Quantitative average results show that FIESTA yielded the best Dice, voxel-wise Dice, voxel-wise bundle adjacency, and streamline density correlation with 0.74$\pm$0.08, 0.96$\pm$0.04, 0.43$\pm$0.21 mm, and 0.85$\pm$0.13 respectively, whereas \textcolor{black}{FINTA-multibundle} obtained the best streamline-wise bundle adjacency with a value of 5.02$\pm$0.54 mm. The voxel-wise bundle adjacency shows that, on average, two binary masks of the same bundle from the same subject are $\sim$0.5 mm apart using FIESTA. For each metric where FIESTA did not come up first, it was in the margin of error of the first one. \textit{TractSeg} yielded the best ICC length and ICC volume with values of with 0.79$\pm$0.05 and 0.82$\pm$0.06, respectively, whereas FIESTA yielded 0.79$\pm$0.05 and 0.81$\pm$0.07, respectively. \textcolor{black}{FINTA-multibundle} yielded the best streamline-wise bundle adjacency (5.02$\pm$0.54 mm), whilst FIESTA yielded 5.24$\pm$0.44 mm. Finally, according to Koo and Li (2016) \cite{koo_guideline_2016}, ICCs yielded by FIESTA show that the method has a good reliability.

\begin{table}[hb!]
\centering
\caption{Average ($\pm$SD) similarity metric values and ICC for the 17 studied bundles, shared across different state-of-the-art methods for the {\em MyeloInferno} dataset. Up and down arrows mean that higher and lower are better respectively for the studied metric. V. and S. stand for Voxels and Streamlines. L. stands for Length. BA stands for Bundle Adjacency.  The bundle adjacency streamlines and the ICC length were not computed for \textit{XTRACT} because it outputs streamline density maps instead of WM streamlines for each bundle. FINTA-m stands for \textcolor{black}{FINTA-multibundle}.}
\begin{tabular}{|l|ccccccc|}
\hline
\makecell{17 Bundles}               	& \makecell{\textit{TractSeg}} & \makecell{RB} & \makecell{RBx} & \makecell{WMA} & \makecell{\textit{XTRACT}}	& \makecell{FINTA-m} 		  & \makecell{FIESTA}     \\ \hline
Dice V. ($\uparrow$)                    & 0.65$\pm$0.10           	   & 0.56$\pm$0.22 & 0.58$\pm$0.23  & 0.56$\pm$0.11  & 0.64$\pm$0.15 			 	& 0.71$\pm$0.09 			  & \textbf{0.74$\pm$0.08}	\\
W-Dice V. ($\uparrow$)           & 0.85$\pm$0.09           	   & 0.78$\pm$0.29 & 0.79$\pm$0.29  & 0.78$\pm$0.14  & 0.84$\pm$0.14 			 	& 0.94$\pm$0.06 			  & \textbf{0.96$\pm$0.04}\\
B. A. V. (mm) ($\downarrow$) & 0.70$\pm$0.45           	   & 1.65$\pm$2.03 & 1.46$\pm$2.01  & 0.88$\pm$0.39  & 0.89$\pm$0.63 			 	& 0.51$\pm$0.27 			  & \textbf{0.43$\pm$0.21}	\\
B. A. S. (mm) ($\downarrow$) & 5.37$\pm$0.92           	   & 7.82$\pm$3.01 & 6.94$\pm$2.76  & 6.07$\pm$0.65  & N/A			 			 	& \textbf{5.02$\pm$0.54}	  & 5.24$\pm$0.44 \\
Density Corr. ($\uparrow$)        & 0.64$\pm$0.18           	   & 0.60$\pm$0.30 & 0.64$\pm$0.30  & 0.58$\pm$0.22  & 0.63$\pm$0.21 			 	& 0.81$\pm$0.15 			  & \textbf{0.85$\pm$0.13}	 \\
ICC L. ($\uparrow$)                 & \textbf{0.79$\pm$0.05}  	   & 0.55$\pm$0.28 & 0.67$\pm$0.19  & 0.78$\pm$0.08  & N/A           			 	& 0.78$\pm$0.06        		  & \textbf{0.79$\pm$0.06} \\
ICC V. ($\uparrow$)                 & \textbf{0.82$\pm$0.06}  	   & 0.24$\pm$1.54 & 0.70$\pm$0.13  & 0.73$\pm$0.15  & 0.67$\pm$0.17 			 	& 0.80$\pm$0.09 			  & 0.81$\pm$0.07 \\\hline
\end{tabular}
\label{tab:average_overlap_metrics}
\end{table}

\begin{figure}[tp]
    \centering
    \includegraphics[width=1
    \textwidth]{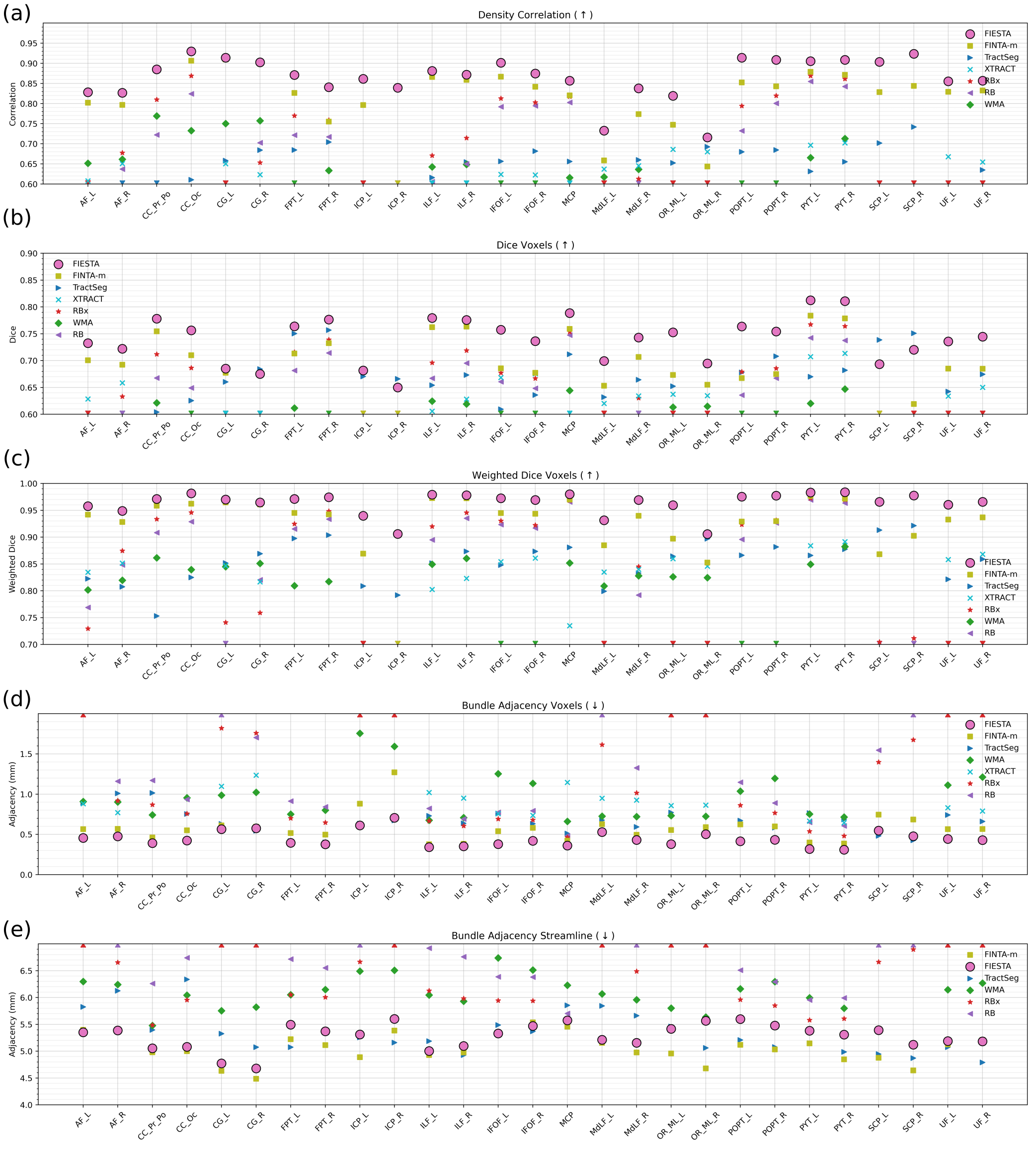}
    \caption{Average per-bundle overlap metrics for the {\em MyeloInferno} dataset.  Each metric was computed over 5 acquisitions from 18 subjects registered to the MNI152 template with per subject pairwise comparison yielding 10 comparisons per subject. For visualization purposes, only the 27 most similar bundles across methods are shown. It should be noted that the \texttt{SCP} was not present in the bundles from WMA. \textit{XTRACT} also misses all \texttt{CC} bundles, the \texttt{FPT}, the \texttt{ICP}, the \texttt{POPT}, and the \texttt{SCP}. (a) Streamline Density Correlation, (b) Voxel-Wise Dice Score Coefficient, (c) Weighted Voxel-Wise Dice Score Coefficient, (d) Voxel-Wise Bundle Adjacency, and (e) Streamline-Wise Bundle Adjacency. The streamline-wise bundle adjacency was not computed for \textit{XTRACT} because the software outputs streamline density volume maps instead of WM streamlines for each bundle. The legend of each chart presents, from top to bottom, the method with the best average metric value according to Table \ref{tab:average_overlap_metrics}. Outlier values are clipped to the bottom or top of each chart and are displayed as small triangles. FINTA-m stands for \textcolor{black}{FINTA-multibundle}. \vspace{-0.5cm}}
    \label{fig:overlap_metrics}
\end{figure}

Fig. \ref{fig:overlap_metrics} presents per-bundle similarity metric results of the 27 benchmarked bundles. It should be noted that \textit{XTRACT} only presents 17 bundles and WMA 25 (the \texttt{SCP} is not present). Except for the streamline-wise bundle adjacency, where \textcolor{black}{FINTA-multibundle} is clearly superior, FIESTA is globally more reliable with better overall scores. The improvement is particularly important over small and hard-to-track bundles such as the \texttt{ICP}, the \texttt{SCP}, the \texttt{OR\_ML} or the \texttt{UF}. More specifically, the streamline density correlation, the Dice score and the weighted Dice score (Fig. \ref{fig:overlap_metrics}a, \ref{fig:overlap_metrics}b, and \ref{fig:overlap_metrics}c), show that FIESTA yields higher within-bundle similarity. Fig. \ref{fig:overlap_metrics}a shows that the streamline density between different acquisitions is systematically more correlated when using FIESTA than with the other methods. The voxel-wise bundle adjacency of FIESTA shows that the average distance between 2 bundle masks of the same subject is $\sim$0.5 mm. Knowing that the voxel size of the \textit{MyeloInferno} dataset is 2 mm isotropic, this underlines how reproducible FIESTA is. 
Finally, Fig. \ref{fig:overlap_metrics}e shows that, depending on the studied bundle, the \textit{TractSeg} streamline-wise bundle adjacency is close to FIESTA and \textcolor{black}{FINTA-multibundle}, with many bundles surpassing the proposed method, whilst never surpassing \textcolor{black}{FINTA-multibundle}.

According to Fig. \ref{fig:overlap_metrics}e, \textcolor{black}{FINTA-multibundle} shows the closest scores to FIESTA with \textit{RecoBundles}, \textit{RecoBundlesX}, and WMA's being less reproducible. In comparison, \textit{RecoBundles}, \textit{RecoBundlesX}, \textcolor{black}{FINTA-multibundle}, and FIESTA had about \num[group-separator={,}]{4000000} input streamlines. This might explain why WMA seems to perform poorly, as we had to down sample the input whole-brain tractogram due to the computational cost of the method.

\begin{figure}[tp]
    \centering
    \includegraphics[width=1
    \textwidth]{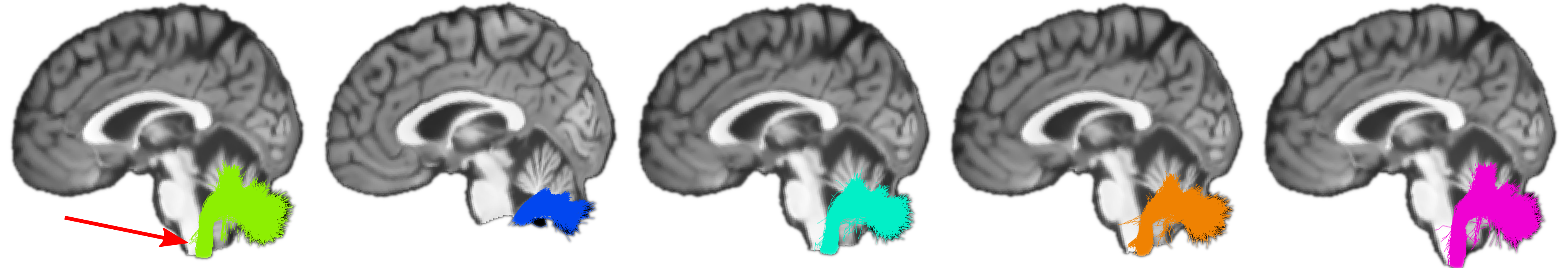}
    \caption{Visualization of FIESTA's \texttt{ICP} across 5 acquisitions of the subject with the largest variability in scores. Each color stands for a different acquisition. The variability is mostly explained by the MRI field of view, and not by FIESTA's natural variability. In that case, the \texttt{ICP} starts at various points of the medulla (or the pons for the second image) because the original image is cut at various points in the brainstem. Red arrows point to the medulla on the left most image.}
    \label{fig:icp_overlap}
\end{figure}

According to the charts in Fig. \ref{fig:overlap_metrics}b and Fig. \ref{fig:overlap_metrics}c, FIESTA is less effective for the \texttt{ICP} bundle. To better understand this behavior, we further investigated the case. Fig. \ref{fig:icp_overlap} shows the left \texttt{ICP} produced by FIESTA on the subject with the poorest similarity metrics. We see that the variability for this particular bundle is explained by the positioning of the MRI field of view, which seems to cut at various points in the brainstem. Therefore, the starting zones of the \texttt{ICP} are variable. Despite this bundle variability, Fig. \ref{fig:icp_overlap} shows that FIESTA is robust to such starting zone variations. 

\begin{figure}[ht]
    \centering
    \includegraphics[width=1
    \textwidth]{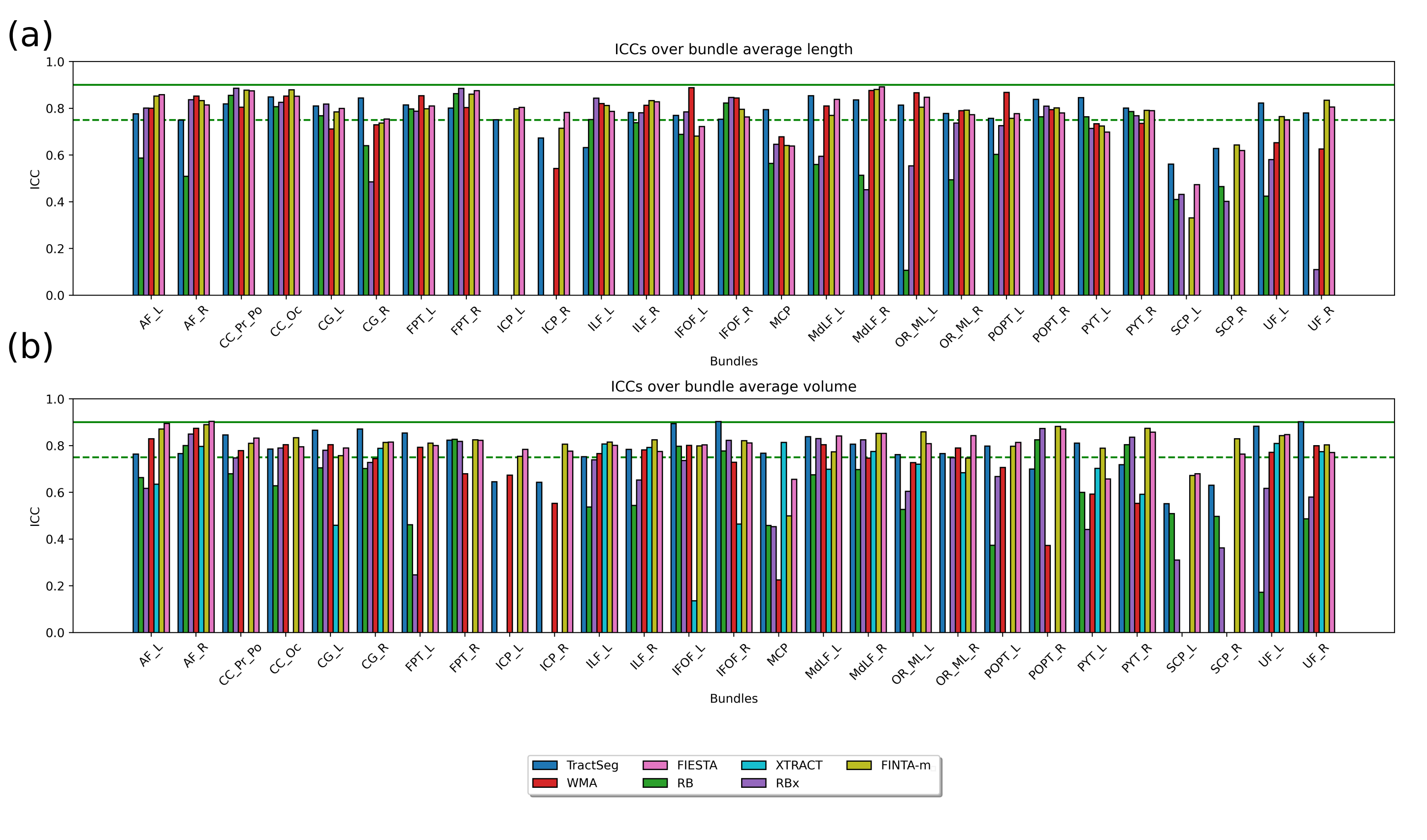}
    \caption{Bundle ICCs from the on the 27 benchmarked bundles, where (a) is the ICC based on bundle average length and (b) is the ICC based on the per-bundle voxel volume. It should be noted that the ICC length was not computed for \textit{XTRACT} because the software outputs streamline density volume maps instead of WM streamlines for each bundle. Green dotted lines indicate good reliability (0.75) and green hard lines indicate excellent reliability (0.9). FINTA-m stands for \textcolor{black}{FINTA-multibundle}.}
    \label{fig:iccs}
\end{figure}

To complement the previous results, Fig. \ref{fig:iccs} shows the reliability scores with respect to bundle-wise metrics – namely volume and average length. Following Koo and Li (2016) \cite{koo_guideline_2016} guidelines, an ICC between 0.75 and 0.9 and above 0.9 are respectively indicative of good and excellent reliability. Fig. \ref{fig:iccs}a presents the ICC of the bundle average length, which provides insight into the reliability of each method to obtain the average bundle length values. FIESTA outperforms \textit{RecoBundles} and \textit{RecoBundlesX} in most bundles. WMA and \textit{TractSeg} are more competitive than anticipated. Careful analysis reveals that although average overlap metrics for those methods are much worse than FIESTA, they yield streamlines with consistent lengths across various time points. 
It can be seen in Fig. \ref{fig:overlap_metrics}e, that \textit{TractSeg} seems to produce a high streamline-wise bundle adjacency. 
The per-bundle volume measurement results in Fig. \ref{fig:iccs}b shows a similar trend with \textit{TractSeg}, \textcolor{black}{FINTA-multibundle}, and FIESTA having comparable results, and \textit{TractSeg} and \textcolor{black}{FINTA-multibundle} outperforming FIESTA for some bundles. This might be explained by the fact that \textit{TractSeg} uses probabilistic bundle specific tracking instead of whole-brain tracking. 
But although the \textit{TractSeg} bundle-wise volume is stable through different time-points, its test-retest similarity metrics are much worse than FIESTA.  
Finally, even if some methods seem to yield higher ICC values than FIESTA, most FIESTA's ICC were higher than 0.75, thus producing good reliability measurement (c.f. Table \ref{tab:average_overlap_metrics}).

\begin{figure}[tp]
    \centering
    \includegraphics[width=1
    \textwidth]{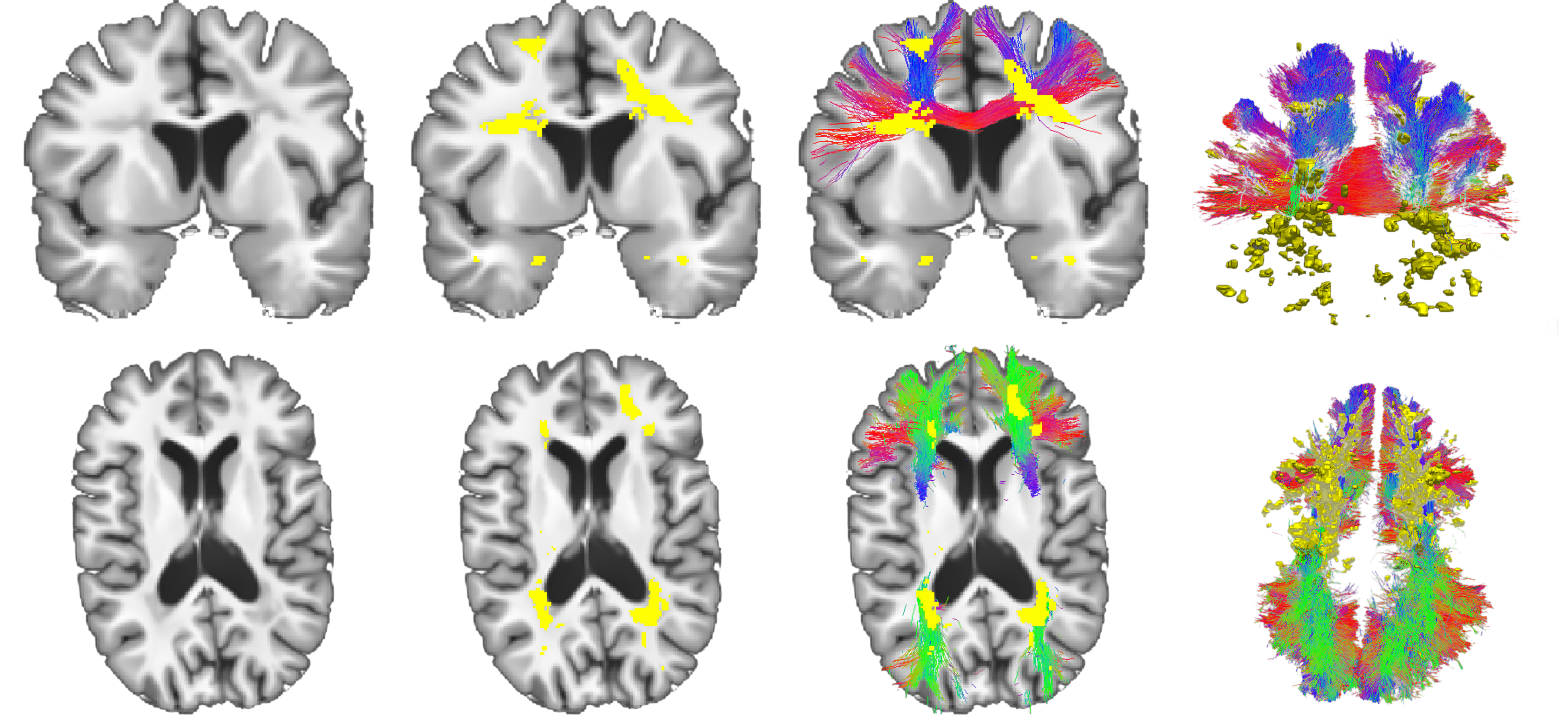}
    \caption{\textcolor{black}{Lesions from one subject in the \textit{MyeloInferno-MS-TR} dataset; \textbf{Top}: Lesions in the \texttt{CC\_Pr\_Po} bundle; \textbf{Bottom}: Lesions in the left and right \texttt{IFOF} bundles.}}
    \label{fig:ms_lesion_bundles}
\end{figure}

\textcolor{black}{Table \ref{tab:disease_average_overlap_metrics} presents the results of FIESTA on three disease cohorts -- namely ADNI-TR, PPMI, and \textit{MyeloInferno-MS-TR}. Results show that performances on those cohorts are similar to what were obtained on healthy subjects, presented in Table \ref{tab:average_overlap_metrics}. Thus, it shows the potential of using FIESTA on a wide range of population. Also, results presented in Table \ref{tab:disease_average_overlap_metrics} shows the usefulness of using \textcolor{black}{GESTA-gmm} to improve \textcolor{black}{most} reliability metrics. Finally, a subset of bundles (\texttt{CC\_Pr\_Po} and \texttt{IFOF}) alongside the segmented lesions are presented in Fig. \ref{fig:ms_lesion_bundles} from one subject of the \textit{MyeloInferno-MS-TR} dataset showing how the bundle are able to cross MS lesions as long as the diffusion signal is not altered in those regions.}

\begin{table}[hb!]
\caption{\textcolor{black}{Results of the performance of FIESTA on three disease cohorts – namely ADNI-TR (AD), PPMI (PD), and MyeloInferno-MS-TR (MS) on the 27 benchmarked bundles; \textbf{Left}: Reliability results of the \textcolor{black}{FINTA-multibundle} module; \textbf{Right}: Reliability results of the FIESTA segmented bundles. FINTA-m stands for FINTA-multibundle.} }
\centering
\begin{tabular}{|l|cc|cc|cc|}
\hline
\makecell{27 bundles}			 & \makecell{AD FINTA-m}        & \makecell{AD FIESTA} 			& \makecell{PD FINTA-m}   	& \makecell{PD FIESTA}  		& \makecell{MS FINTA-m}   	& \makecell{MS FIESTA}  			\\ \hline
Dice V. ($\uparrow$)             & 0.69$\pm$0.10			 	 & \textbf{0.73$\pm$0.08}         	& 	0.68$\pm$0.09			 	&   \textbf{0.74$\pm$0.07}		  	& 	0.70$\pm$0.10			 	& \textbf{0.76$\pm$0.05	}				\\
W-Dice V. ($\uparrow$)           & 0.91$\pm$0.12			 	 & \textbf{0.95$\pm$0.09} 	    	& 	0.91$\pm$0.10			 	&   \textbf{0.96$\pm$0.06}		  	& 	0.93$\pm$0.09			 	& \textbf{0.97$\pm$0.04}					\\
B. A. V. (mm) ($\downarrow$) 	 & 0.48$\pm$0.23			 	 & \textbf{0.43$\pm$0.21}         	& 	0.53$\pm$0.29			 	&   \textbf{0.40$\pm$0.23}		  	& 	0.49$\pm$0.30			 	& \textbf{0.37$\pm$0.14}					\\
B. A. S. (mm) ($\downarrow$) 	 & \textbf{4.63$\pm$0.51}			 	 & 5.40$\pm$0.43         	& 	\textbf{5.35$\pm$0.48}			 	&   5.35$\pm$0.49		  	& 	\textbf{5.23$\pm$0.48}			 	& 5.27$\pm$0.38					\\
Density Corr. ($\uparrow$)       & 0.82$\pm$0.19			     & \textbf{0.90$\pm$0.14}			& 	0.83$\pm$0.16				&   \textbf{0.91$\pm$0.11}	       & 	0.87$\pm$0.12				& \textbf{0.93$\pm$0.08}					\\
ICC L. ($\uparrow$)              & \textbf{0.95$\pm$0.03}	     & 0.94$\pm$0.03			       	& 	\textbf{0.92$\pm$0.03}				& \textbf{0.92$\pm$0.03}            & 	0.92$\pm$0.08				& \textbf{0.94$\pm$0.05}					       		\\
ICC V. ($\uparrow$)              & 0.95$\pm$0.04				 & \textbf{0.96$\pm$0.02}			& 	0.90$\pm$0.06			 	&  \textbf{0.91$\pm$0.05} 		  	& 	0.92$\pm$0.10			 	& \textbf{0.94$\pm$0.08}							    \\
\hline
\end{tabular}
\label{tab:disease_average_overlap_metrics}
\end{table}

\begin{figure}[ht]
    \centering
    \includegraphics[width=1
    \textwidth]{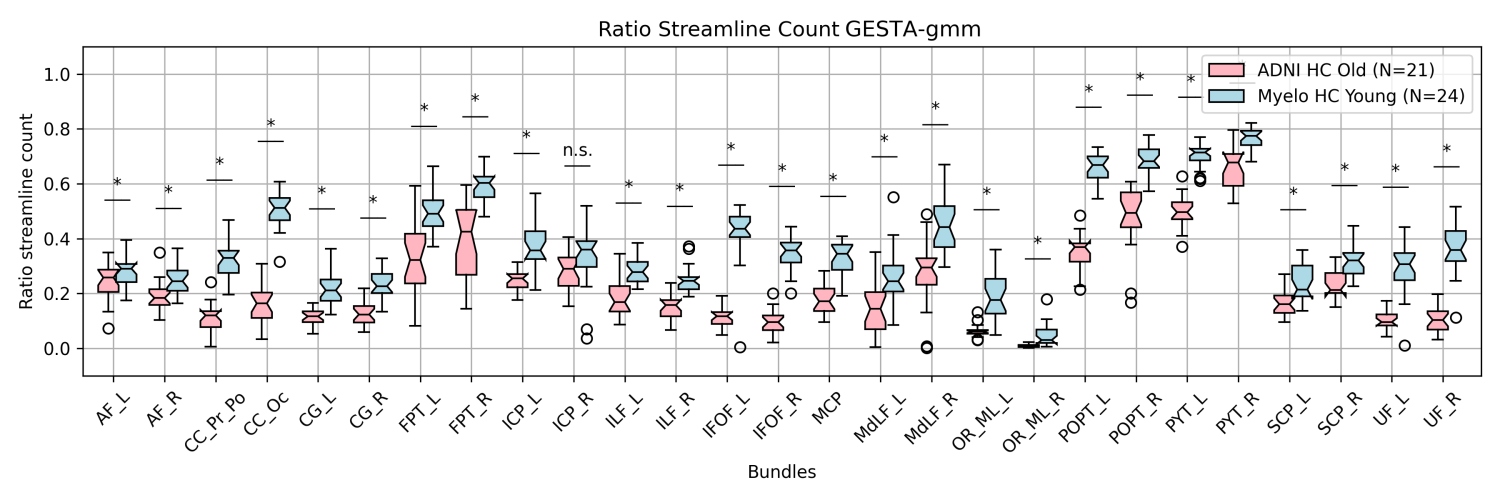}
    \caption{\textcolor{black}{Ratio between the streamline count for each bundle before and after the four anatomical constraint filtering. * indicates a $p$ value < 0.05, n.s. means non statistically significant difference}}
    \label{fig:r_streamline_count}
\end{figure}

\textcolor{black}{\subsection{\textcolor{black}{GESTA-gmm} generated streamlines faithfulness analysis}
Fig. \ref{fig:r_streamline_count} presents the results showing the ratio between the streamline count for each bundle of the generated streamlines ($N=\num[group-separator={,}]{25000}$) before the four anatomical constraints filtering and after. Thus, a ratio of 0.5 indicates that, for a particular bundle, for each remaining streamline at the end of the filtering process, 2 must be generated by the rejection sampling process. Fig. \ref{fig:r_streamline_count} shows that our initial hypothesis,  stipulating that a greater percentage of generated streamlines should be accepted after passing the
4 constraints on young subjects, explains such results. Indeed, for all bundles, more streamlines passed the filtering process for the young population in comparison to the old population. Such results suggest that the remaining streamlines are faithful to the underlying anatomy and respect the diffusion signal for both the young and old subjects.}

\textcolor{black}{Fig. \ref{fig:bas_adni_myelo} shows the results of the streamline-wise bundle adjacency between the young subjects and the old subjects for each bundle. We see that, as anticipated, the metric improves with the usage of \textcolor{black}{GESTA-gmm}. On average, the streamline-wise bundle adjacency for FINTA-multibundle was 6.08$\pm$0.65 mm, whilst it was 5.98$\pm$0.48 mm when using FIESTA. However, such a small augmentation of 0.1 mm, in comparison to the smallest ADNI voxel size (1.36 mm), indicates that the generated streamlines that pass the 4 anatomical constraints, even if the atlas bundles and anatomy were based on a younger population, stay faithful to the current subject anatomy.}

\textcolor{black}{\begin{figure}[ht]
    \centering
    \includegraphics[width=1
    \textwidth]{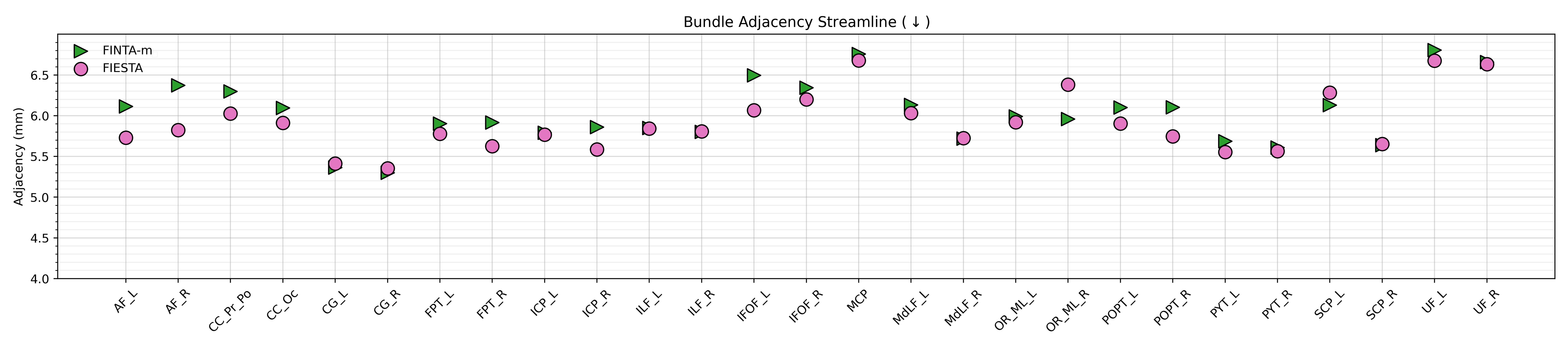}
    \caption{\textcolor{black}{Streamline-wise, bundle adjacency for the FIESTA and the FINTA-multibundle bundles between the young and the old healthy subjects from the \textit{MyeloInferno-HC} and ADNI-HC datasets. FINTA-m stands for FINTA-multibundle.}}
    \label{fig:bas_adni_myelo}
\end{figure}}

\section{Discussion}
\label{sec:discussion}

Automatic \textcolor{black}{bundle segmentation} is an essential processing step in tractography to isolate specific white matter pathways. In this study, we presented a novel semi-supervised autoencoder-based automatic \textcolor{black}{bundle segmentation} method called FIESTA (\textcolor{black}{\textit{FIbEr Segmentation in Tractography using Autoencoders}}). We compared FIESTA's reliability to state-of-the-art automatic \textcolor{black}{bundle segmentation} methods -- namely \textit{RecoBundles} (RB) \cite{garyfallidis_recognition_2018}, \textit{RecoBundlesX} (RBx) \cite{rheault_analyse_2020}, \textit{TractSeg} \cite{wasserthal_tract_2018, wasserthal_tractseg_2018, wasserthal_combined_2019, wasserthal_multiparametric_2020}, \textit{XTRACT} \cite{warrington_xtract_2020}, and \textit{White Matter Analysis} (WMA) \cite{odonnell_automatic_2007, odonnell_unbiased_2012, zhang_anatomically_2018}. We found that methods such as \textit{RecoBundles} and \textit{RecoBundlesX} are prone to large intra and inter-subject variability. Bundle definitions from \textit{TractSeg} are fixed, requiring a complete retraining of the system for a different set of bundles. From a practical point of view, it would be hardly doable to change a specific bundle definition overnight, as one would need to create a completely new training set to train all three U-Nets. WMA is too computationally expensive. We had to downsample input tractograms to \num[group-separator={,}]{500000} streamlines. \textit{XTRACT} ROIs seem to yield bundles with a low specificity and are hardly editable because they are defined by 5 anatomical experts, which would be time-consuming to change. Also, we found that FIESTA is robust to these problems present in current state-of-the-art \textcolor{black}{bundle segmentation} methods. \textcolor{black}{Moreover, we showed that FIESTA is robust and reliable on various kinds of brain populations, such as disease cohorts, as long as the diffusion signal is not altered (or slightly altered) by such diseases.}

As mentioned, the main goal of this work is to assess the reliability of FIESTA alongside state-of-the-art automatic \textcolor{black}{bundle segmentation} methods. As per many segmentation algorithms in biomedical image analysis, we claim that reliability analysis are as important as standard comparisons to some ground truth (or silver standard). It gives a better insight on the robustness of the method, even if the comparison to the ground truth is poor. Therefore, both validation methods are as important, but the reliability is often rarely reported. Results presented in Table \ref{tab:average_overlap_metrics} show that FIESTA yields reliable bundles suggested by different similarity metrics (Dice, weighted-Dice, voxel-wise bundle adjacency, streamline-wise bundle adjacency and streamline density correlation) in a test-retest manner over different dMRI time points. Quantitative average results showed that FIESTA yielded the best Dice, voxel-wise bundle adjacency, and streamline density correlation or is in the margin of error of the best values (streamline-wise bundle adjacency, ICC length, and ICC volume). The voxel-wise bundle adjacency shows that, on average, two binary masks of the same bundle from the same subject are $\sim$0.5 mm apart using FIESTA. This value is minimal considering that the acquisition resolution of the dataset used is 2 mm isotropic. Also, variability induced by registration, necessary for similarity metric computations, did not overly affect the voxel-wise bundle adjacency average distance. Higher density correlations obtained by FIESTA might be explained, in part, by the use of generative sampling. Interestingly, a closer analysis of the source of variability in FIESTA reveals that it is mainly due to the MRI field of view positioning (c.f. Fig. \ref{fig:icp_overlap}).

ICC results are not as consistent as it is the case for similarity metrics. It is seen that bundles from \textit{TractSeg} are more reliable for volume and average length. This is probably due to the fact that \textit{TractSeg} uses probabilistic bundle-specific tracking and not whole-brain tracking for bundle \textcolor{black}{segmentation} later on. Thus, it probably yields more consistent metrics at various time points depending on the studied bundle. Following those results, we claim that the interpretation of any ICCs, especially in biomedical image analysis, needs to be interpreted cautiously. Indeed, a bad algorithm could, for example, produce, by chance, bundles with the same volume without being in the same anatomical region. Thus, it would yield high ICCs, with low overlap values. Therefore, the interpretability of an ICC must always be paired, within the context of a tractography analysis, with similarity metrics and a good qualitative analysis. 

\textcolor{black}{The analysis of the generative sampling scheme shows that the generated streamlines remaining after the filtering process respect the subject anatomy, and do not generalize to a generic ensemble of streamlines (c.f. Fig. \ref{fig:r_streamline_count} and \ref{fig:bas_adni_myelo}). Such behavior could be assumed to happen due to the fact that part of the latent space is seeded using generic streamlines based on the PAWM atlas. However, the constraint on the diffusion signal removes all streamlines that are not faithful to one's subject anatomy. Such constraint stipulates that any streamline generated using the generative process that has a local orientation angle to fODF peak greater than a certain angle value (i.e., 30\degree) ought to be discarded. Therefore, we would argue that streamlines that are kept after this step should be closely similar to a streamline generated from a traditional tractography process. Another way to see this filtering step is that, instead of tracking the streamline on the diffusion signal, we proposed streamlines that are likely to be close to the subject anatomy, but we only keep the ones that respect the diffusion signal.}

FIESTA is not a method without flaws. First, we see that the use of generative sampling does worsen the streamline-wise bundle adjacency. We hypothesized that such behavior might be due to the usage of atlas bundles as seeds for the latent space \textcolor{black}{combined with the lossy process of autoencoders. Thus, combining generative sampling with generic seed points and another kind of tractography process might produce variations that increase the streamline-wise bundle adjacency.} Such behavior is currently under investigation, and generative sampling filtering parameters might need further adjustment. Second, the threshold calibration, even though it can be done quickly compared to other state-of-the-art methods, is probably the most time-consuming process. A solution would be to get rid of the threshold method by including a new \textit{bundle} in the atlas with all whole-brain implausible WM streamlines. Therefore, streamlines closer to the implausible \textit{bundle} would be classified as implausible. New bundle definitions could easily be implemented, as only an ideal bundle in a standard space would be needed for the change to be effective, without the need to find/calibrate any threshold. As WMA uses this method for implausible streamlines, it might be pertinent to test. Unfortunately, this method also has its limitations, as defining implausible streamlines is an open and hard problem to solve. Also, even if we presented a solution to find the \textit{\textcolor{black}{near-}optimal} threshold in section \ref{sec:threshold_calibration}, the most easy way to find the desired threshold, whilst being the least sophisticated method, is just to test a few thresholds manually, starting completely randomly and slowly converging around the desired final values. Moreover, the generative sampling method takes a long time for certain bundles. It would be interesting to include per-bundle generative sampling parameters to optimize the pipeline \cite{legarreta_generative_2022}. \textcolor{black}{Furthermore, FIESTA was not developed to work properly on heavily impaired and deformed brain anatomy (such as tumor). Such, behavior was not characterized, and we are currently investigating the best approach for such specific brain deformed anatomy. However, we showed that FIESTA is robust to small white matter lesions as long as the diffusion signal is not (or slightly) altered. Also, we showed that the brain deformation on normal aging population does not affect the performances of FIESTA. Finally, we can see in Table \ref{tab:method_params} that GESTA-gmm is still a slow process, even though it was improved for this pipeline. Thus, more optimization work is needed for the generative process. Even if a method like \textit{TractSeg} is still faster than FIESTA, its rigid framework prevents rapid iterations when new bundles are needed. Here, we demonstrated that it is possible to have a bundle segmentation pipeline that works on raw tractograms and is more reliable in a test-retest manner than state-of-the-art bundle segmentation methods.}

This work mainly focuses on the \textcolor{black}{bundle segmentation} of long-range white matter streamlines, that connect distant cortical areas of the brain. We did not assess the behavior of FIESTA on short association fibers (U-Fibers) \cite{shastin_surface-based_2022} that connect adjacent cortical regions. Future works will include the evaluation of FIESTA on such fibers. 

\section{Conclusion}
\label{sec:conclusion}
We presented a new tractography \textcolor{black}{bundle segmentation} pipeline named FIESTA (\textcolor{black}{\textit{FIbEr Segmentation in Tractography using Autoencoders}}) that leverages the power of unsupervised and self-supervised learning and that is intended as a solution to the limitations of current state-of-the-art \textcolor{black}{bundle segmentation} pipelines. We showed that FIESTA lets users easily and rapidly edit its bundle definitions, and is highly reliable in test-retest. FIESTA is more reliable than other state-of-the-art methods such as \textit{TractSeg}, \textit{RecoBundles}, \textit{RecoBundlesX}, \textit{XTRACT}, and WMA. We believe FIESTA reaches an optimal compromise between computational burden, ease-of-use, and reconstruction quality and reliability. Thus, FIESTA might be a promising avenue as an easy and reliable way for \textcolor{black}{bundle segmentation} in tractography, especially so for hard-to-track bundles.

\newpage
\section*{Data/code availability}

All data used to develop and test FIESTA were acquired from different public (TractoInferno, HCP, PAWM atlas\textcolor{black}{, ADNI, PPMI}) and private datasets (\textit{MyeloInferno}). Data from the MyeloInferno dataset cannot be made public to respect ethics protocol of the institution. \textcolor{black}{FIESTA config files, training data, and the deep learning model are made publicly available at \url{https://doi.org/10.5281/zenodo.7562790} \cite{dumais_felix_2023_7562790}}. URL links for public datasets are:

\begin{enumerate}
    \item \textit{TractoInferno}:  \url{https://openneuro.org/datasets/ds003900}
    \item HCP: \url{http://www.humanconnectomeproject.org/}
    \item PAWM atlas: \url{https://doi.org/10.5281/zenodo.5165374}
    \item \textcolor{black}{ADNI: \url{https://adni.loni.usc.edu/}}
    \item \textcolor{black}{PPMI: \url{https://www.ppmi-info.org/}}
\end{enumerate}

\textbf{The source code will be made available to the community upon the acceptance of the article under the \url{https://github.com/scil-vital} GitHub organization.}

\section*{Ethics statement}
Written informed consent was obtained from participants and were recruited following the ethics protocol of the Centre de Recherche du Centre Hospitalier Universitaire de Sherbrooke (Sherbrooke, Canada) (for the MyeloInferno dataset).

\section*{Acknowledgments}
This research was conducted in the context of the Acuity-QC group funded by the Fond d'Accélération des Collaborations en Santé (FACS) and Consortium Québecois sur la Découverte du Médicament (CQDM). It was enabled, in part, by support provided by Calcul Québec (www.calculquebec.ca) and Digital Research Alliance of Canada (www.alliancecan.ca).

Data were provided, in part, by the Human Connectome Project, WU-Minn Consortium (Principal Investigators: David Van Essen and Kamil Ugurbil; 1U54MH091657) funded by the 16 NIH Institutes and Centers that support the NIH Blueprint for Neuroscience Research; and by the McDonnell Center for Systems Neuroscience at Washington University.

Data were provided, in part, by the OpenNeuro database (\url{https://openneuro.org/datasets/ds003900}). Its accession number is ds003900.

\textcolor{black}{Data collection and sharing for this project was funded, in part, by the Alzheimer's Disease Neuroimaging Initiative (ADNI) (National Institutes of Health Grant U01 AG024904) and DOD ADNI (Department of Defense award number W81XWH-12-2-0012). ADNI is funded by the National Institute on Aging, the National Institute of Biomedical Imaging and Bioengineering, and through generous contributions from the following: AbbVie, Alzheimer's Association; Alzheimer's Drug Discovery Foundation; Araclon Biotech; BioClinica, Inc.; Biogen; Bristol-Myers Squibb Company; CereSpir, Inc.; Cogstate; Eisai Inc.; Elan Pharmaceuticals, Inc.; Eli Lilly and Company; EuroImmun; F. Hoffmann-La Roche Ltd and its affiliated company Genentech, Inc.; Fujirebio; GE Healthcare; IXICO Ltd.; Janssen Alzheimer Immunotherapy Research \& Development, LLC.; Johnson \& Johnson Pharmaceutical Research \& Development LLC.; Lumosity; Lundbeck; Merck \& Co., Inc.; Meso Scale Diagnostics, LLC.; NeuroRx Research; Neurotrack Technologies; Novartis Pharmaceuticals Corporation; Pfizer Inc.; Piramal Imaging; Servier; Takeda Pharmaceutical Company; and Transition Therapeutics. The Canadian Institutes of Health Research is providing funds to support ADNI clinical sites in Canada. Private sector contributions are facilitated by the Foundation for the National Institutes of Health (www.fnih.org). The grantee organization is the Northern California Institute for Research and Education, and the study is coordinated by the Alzheimer's Therapeutic Research Institute at the University of Southern California. ADNI data are disseminated by the Laboratory for Neuro Imaging at the University of Southern California.}

\textcolor{black}{Data used in the preparation of this article were obtained from the Parkinson's Progression Markers Initiative (PPMI) database (www.ppmi-info.org/access-data-specimens/download-data). For up-to-date information on the study, visit \url{www.ppmi-info.org}. PPMI – a public-private partnership – is funded by the Michael J. Fox Foundation for Parkinson's Research and funding partners, including [list the full names of all of the PPMI funding partners found on the \url{https://www.ppmi-info.org/about-ppmi/who-we-are/study-sponsors}].}

\section*{Conflicts of Interest} Maxime Descoteaux and Pierre-Marc Jodoin report membership and employment with Imeka Solutions inc. Patent \#17/337,413 is pending to Imeka Solutions inc. with inventors Jon Haitz Legarreta, Maxime Descoteaux and Pierre-Marc Jodoin. Muhamed Barakovic is an employee of Hays plc and a consultant for F. Hoffmann-La Roche Ltd. Stefano Magon is an employee and shareholder of F. Hoffmann-La Roche Ltd.

\newpage
\bibliographystyle{abbrvnat}  
\bibliography{references}  

\newpage
\appendix
\section{Appendices}
\setcounter{table}{0}
\setcounter{figure}{0}
\renewcommand\thetable{\Alph{section}.\arabic{table}}
\renewcommand\thefigure{\Alph{section}.\arabic{figure}}

\label{sec:appendices}

\subsection{Bundles used for experiments}
\label{sec:bundles}

\begin{table}[h]
\caption{Original 39 bundles provided in the PAWM atlas \cite{rheault_population_2021}. Labels with only one number indicate that there is no lateral symmetry for this particular bundle.}
\centering
\begin{tabular}{llcc}
\hline
\textbf{Name}                                       & \textbf{Abbreviation} & \textbf{Left} & \textbf{Right} \\ \hline
Anterior Commissure                                 & AC                    & \multicolumn{2}{c}{1}          \\
Arcuate Fasciculus                                  & AF                    & 2             & 3              \\
Corpus callosum, Frontal Lobe (most anterior part)  & CC\_Fr\_1             & \multicolumn{2}{c}{4}          \\
Corpus callosum, Frontal Lobe (most posterior part) & CC\_Fr\_2             & \multicolumn{2}{c}{5}          \\
Corpus callosum, Occipital Lobe                     & CC\_Oc                & \multicolumn{2}{c}{6}          \\
Corpus callosum, Parietal Lobe                      & CC\_Pa                & \multicolumn{2}{c}{7}          \\
Corpus callosum, Pre- / Post-Central Gyri              & CC\_Pr\_Po            & \multicolumn{2}{c}{8}          \\
Corpus callosum, Temporal Lobe                      & CC\_Te                & \multicolumn{2}{c}{9}          \\
Cingulum                                            & CG                    & 10            & 11             \\
Frontal Aslant Tract                                & FAT                   & 12            & 13             \\
Fronto-Pontine Tract                                & FPT                   & 14            & 15             \\
Fornix                                              & FX                    & 16            & 17             \\
Inferior cerebellar peduncle                        & ICP                   & 18            & 19             \\
Inferior Fronto-Occipital Fasciculus                & IFOF                  & 20            & 21             \\
Inferior Longitudinal Fasciculus                    & ILF                   & 22            & 23             \\
Middle Cerebellar Peduncle                          & MCP                   & \multicolumn{2}{c}{24}         \\
Middle Longitudinal Fasciculus                      & MdLF                  & 25            & 26             \\
Optic Radiation and Meyer's loop                    & OR\_ML                & 27            & 28             \\
Posterior Commissure                                 & PC                    & \multicolumn{2}{c}{29}         \\
Parieto-Occipito Pontine Tract                      & POPT                  & 30            & 31             \\
Pyramidal Tract                                     & PYT                   & 32            & 33             \\
Superior Cerebellar Peduncle                        & SCP                   & 34            & 35             \\
Superior Longitudinal Fasciculus                    & SLF                   & 36            & 37             \\
Uncinate Fasciculus                                 & UF                    & 38            & 39             \\ \hline
\end{tabular}
\label{tab:bundles}
\end{table}

\subsection{\textcolor{black}{GESTA-gmm} filtering parameters}
\label{sec:dmri_filters}
Table \ref{tab:dmri_filters} presents criteria and threshold values used to ensure that the generated streamlines comply with the dMRI underlying signal and general anatomical constraints. Those were based on the ones used in GESTA \cite{legarreta_generative_2022}. All streamlines outside the length range used are removed. The winding angle is the maximum turning angle that a streamline can make before being removed. The local orientation angle to fODF (LOA-to-fODF) peak and rate values correspond to the minimum number of streamline local orientation vectors where the angle is lower than 30$\degree$ with the closest fODF peak related to the total number of local orientation vectors in the streamline. If the ratio is smaller than 0.75, the streamline is removed. Finally, the WM coverage rate is the percentage of streamline vertices mapping to a WM voxel. 

\begin{table}[h]
\centering
\caption{\textcolor{black}{GESTA-gmm} filtering parameters used after the rejection sampling}
\label{tab:dmri_filters}
\begin{tabular}{lc}
\hline
\textbf{Filter}              & \textbf{Value}     \\ \hline
Length range (mm)            & 20-220             \\
Winding ($\degree$)                   & <360     \\
LOA-to-fODF peak ($\degree$) and rate & <30/0.75 \\
WM coverage rate             & >0.95               \\ \hline
\end{tabular}
\end{table}

\newpage
\subsection{\textcolor{black}{GESTA-gmm} Ratio}
\label{sec:gesta_ratio}
\begin{figure}[!ht]
    \centering
    \includegraphics[width=1\textwidth]{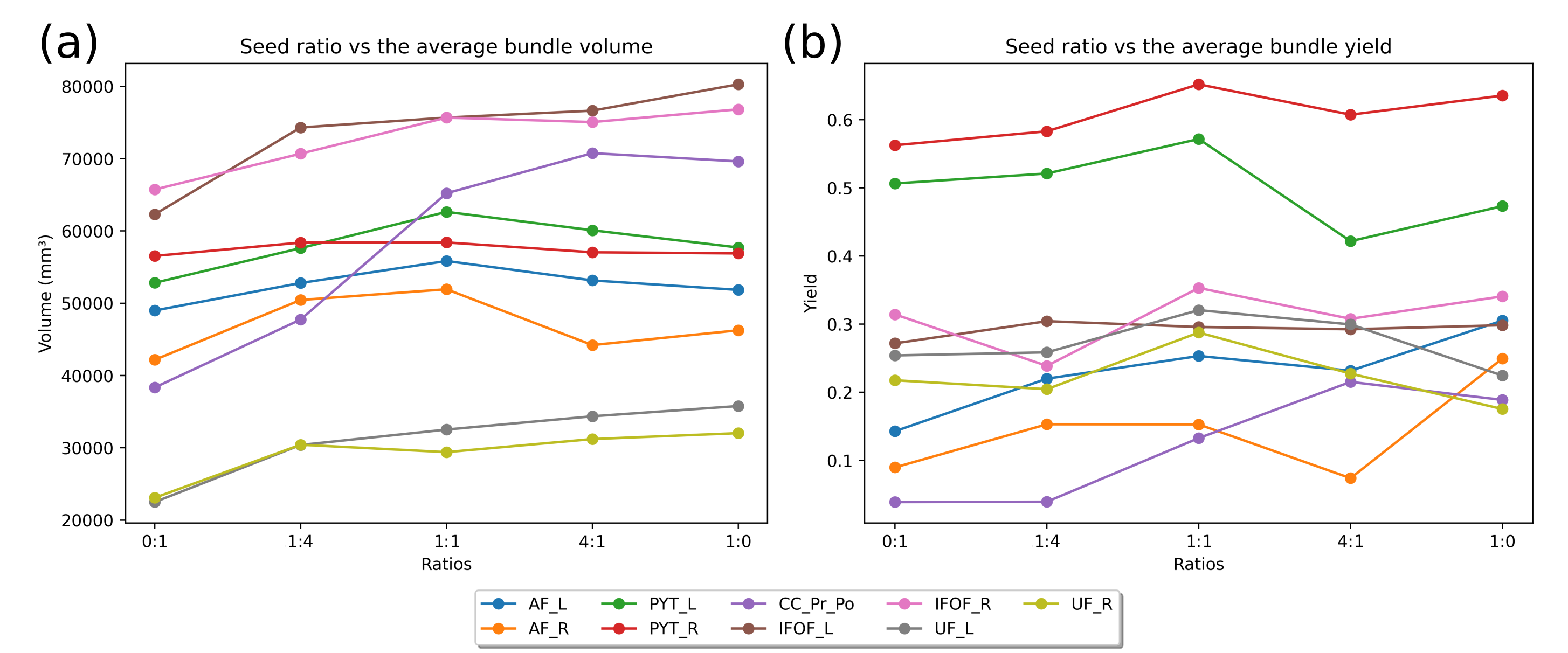}
    \caption{(a) Volume over 9 selected bundles in the PAWM atlas resulting from the \textcolor{black}{GESTA-gmm} process with \num[group-separator={,}]{30000} streamlines sampled in the latent space followed by all the filtering steps in the \textcolor{black}{GESTA-gmm} filtering process. Thus, the final number of streamlines is variable and below \num[group-separator={,}]{30000}. We hereby report volumes post-filtering. (b) Percentage of kept streamlines after the \textcolor{black}{GESTA-gmm} filtering process. The x-axis shows the composition of the latent space, with \textbf{X:Y} being the proportion of the subject bundle streamlines and the proportion of the atlas bundle streamlines.}
    \label{fig:ratio_volume}
\end{figure}

Using \num[group-separator={,}]{10000} streamlines as an empirically determined seed number for the Parzen window, we analyzed the effect of the ratio between the number of atlas bundle seeds and the number of subject bundle seeds. To do so, the data of an HCP subject was processed with the \textit{TractoFlow}~\cite{theaud_tractoflow_2020} pipeline and used for the subject seeds. We tested five subject/atlas bundle ratio values $\textcolor{black}{\{}0{:}1, \textcolor{black}{1}{:}\textcolor{black}{4}, \textcolor{black}{1}{:}\textcolor{black}{1}, \textcolor{black}{4}{:}\textcolor{black}{1}, 1{:}0\textcolor{black}{\}}$. Following those ratios, rejection sampling was used to sample \num[group-separator={,}]{30000} new points over the empirically estimated PDF. Finally, after decoding those new latent points, we filtered according to the plausibility assessment criteria used in GESTA \cite{legarreta_generative_2022} for the length range, the WM coverage, the maximum curving angle, and the local streamline orientation to fODF peak angle (c.f. Supplemental section \ref{sec:dmri_filters} for more details). Fig. \ref{fig:ratio_volume}a presents the latent space seed ratio values for 9 selected bundles -- namely the \texttt{AF\_L}, the \texttt{AF\_R}, the \texttt{PYT\_L}, the \texttt{PYT\_R}, the \texttt{CC\_Pr\_Po}, the \texttt{IFOF\_L}, the \texttt{IFOF\_R}, the \texttt{UF\_L}, and the \texttt{UF\_R} --, in comparison with the final, post filtering steps in the \textcolor{black}{GESTA-gmm} filtering process, obtained WM volumes. Fig. \ref{fig:ratio_volume}b shows the post-filtering \textcolor{black}{GESTA-gmm} streamline yield. A higher yield means that fewer streamlines are removed by all filtering steps in the \textcolor{black}{GESTA-gmm} filtering process. Overall, results suggest that a good trade-off ratio is $\textcolor{black}{1}{:}\textcolor{black}{1}$ i.e., a latent space composed of 50\% of streamlines from the subject bundle and 50\% from the atlas bundle has a good yield rate with a high bundle volume. Thus, in this work, we adopted a subject-to-atlas seed ratio value of $\textcolor{black}{1}{:}\textcolor{black}{1}$.

\newpage
\subsection{Bundles compared across the 5 benchmark state-of-the-art methods}
\label{sec:bundle_compared}

\begin{table}[h]
\label{tab:bundle_compared}
\caption{Bundles compared from each benchmarked method. Empty cells mean that the bundle was not defined by the method and was therefore ignored by subsequent computation}
\centering
\begin{tabular}{l|l|l|l}
\hline
\textbf{RB, RBx, FIESTA} & \textbf{TractSeg} & \textbf{WMA}                 & \textbf{XTRACT}              \\ \hline
AF\_L                   & AF\_left          & T\_AF\_left                  & af\_l                        \\
AF\_R                   & AF\_right         & T\_AF\_right                 & af\_r                        \\
CC\_Oc                  & CC\_7             & T\_CC7                       & \textbf{--} \\
CC\_Pr\_Po              & CC\_4             & T\_CC4                       & \textbf{--} \\
CG\_L                   & CG\_left          & T\_CB\_left                  & cbd\_l                       \\
CG\_R                   & CG\_right         & T\_CB\_right                 & cbd\_r                       \\
FPT\_L                  & FPT\_left         & T\_CR-F\_left                & \textbf{--} \\
FPT\_R                  & FPT\_right        & T\_CR-F\_right               & \textbf{--} \\
ICP\_L                  & ICP\_left         & T\_ICP\_left                 & \textbf{--} \\
ICP\_R                  & ICP\_right        & T\_ICP\_right                & \textbf{--} \\
IFOF\_L                 & IFO\_left         & T\_IOFF\_left                & ifo\_l                       \\
IFOF\_R                 & IFO\_right        & T\_IOFF\_right               & ifo\_r                       \\
ILF\_L                  & ILF\_left         & T\_ILF\_left                 & ilf\_l                       \\
ILF\_R                  & ILF\_right        & T\_ILF\_right                & ilf\_r                       \\
MCP                     & MCP               & T\_MCP                       & mcp                          \\
MdLF\_L                 & MLF\_left         & T\_MdLF\_left                & mdlf\_l                      \\
MdLF\_R                 & MLF\_right        & T\_MdLF\_right               & mdlf\_r                      \\
OR\_ML\_L               & OR\_left          & T\_SO\_left                  & or\_l                        \\
OR\_ML\_R               & OR\_right         & T\_SO\_right                 & or\_r                        \\
POPT\_L                 & POPT\_left        & T\_CR-P\_left                & \textbf{--} \\
POPT\_R                 & POPT\_right       & T\_CR-P\_right               & \textbf{--} \\
PYT\_L                  & CST\_left         & T\_CST\_left                 & cst\_l                       \\
PYT\_R                  & CST\_right        & T\_CST\_right                & cst\_r                       \\
SCP\_L                  & SCP\_left         & \textbf{--} & \textbf{--} \\
SCP\_R                  & SCP\_right        & \textbf{--} & \textbf{--} \\
UF\_L                   & UF\_left          & T\_UF\_left                  & uf\_l                        \\
UF\_R                   & UF\_right         & T\_UF\_right                 & uf\_r                        \\ \hline
\end{tabular}
\end{table}

\newpage
\subsection{PAWM atlas}
\label{sec:pawm_atlas_supp}

For the development and the evaluation of the current pipeline, we used a homemade and publicly available Population Average of WM (PAWM) atlas to evaluate our framework \cite{rheault_population_2021}. The PAWM atlas was built in the context of \textit{RecoBundlesX} works by \citet{garyfallidis_recognition_2018, rheault_analyse_2020} based on ExTractor \cite{petit_structural_2022} and well curated by a neuroanatomist to finally have bundles that fit their normative shapes \cite{rheault_population_2021}. The PAWM atlas was created from 20 random UKBioBank \cite{sudlow_uk_2015} subjects' probabilistic tractograms (ensemble tractography from WM seeding and interface seeding with PFT tracking \cite{girard_towards_2014, garyfallidis_dipy_2014}). Tractograms were filtered for implausible streamlines using ExTractor \cite{petit_structural_2022}, and bundles of interest were extracted from sequences of anatomical rules applied only to anatomically plausible streamlines. 

Nonlinear registration using cross-correlation metric between the subjects' T1w images and the MNI152 2009c Nonlinear Symmetric template was performed using Ants \cite{avants_reproducible_2011}. The affine transformation and deformation fields were applied to bundles, and visual quality control confirmed bundles were adequately reconstructed and warped to the template.

Matching bundles were merged to obtain a representative population average for each bundle. Bundles were then manually cleaned using a clustering approach. Bundles were decomposed into clusters using \textit{QuickBundlesX} \cite{garyfallidis_quickbundlesx_2015}. Each cluster was inspected and accepted or discarded to remove outliers (streamlines with aberrant shapes), incomplete streamlines, or extremely noisy clusters. This manual method allowed thorough filtering of implausible streamlines, while preserving sparse/rare, but anatomically valid, streamlines (i.e., fanning streamlines, streamlines in hard-to-track bundles).

The next step consisted of a symmetrization of both hemispheres. To avoid introducing a left/right bias in FIESTA's \textcolor{black}{bundle segmentation} step, the association and projection atlas bundles were mirrored across hemispheres. Using the spatial transformation of the MNI152 template, bundles were flipped on the X-axis and locally registered to their analogous bundles on the other side using a (constrained) rigid Streamlines Local Registration (SLR) \cite{garyfallidis_robust_2015}. This ensured a better final overlap that matched the expected position of each bundle.

Finally, all bundles were downsampled to a maximum of \num[group-separator={,}]{10000} streamlines to reduce computational burden. Streamlines were upsampled, smoothed using a 3D Gaussian kernel applied to the vertices along each streamline, and re-compressed \cite{presseau_new_2015}. This led to a final lightweight, population-averaged, anatomically meaningful atlas of WM bundles. Fig. \ref{fig:atlas} presents a representation of all the bundles in the PAWM atlas. The bundle full names, abbreviations, and labels used in this work are indicated in Table \ref{tab:bundles}.

\newpage
\subsection{\textcolor{black}{Acronyms}}
\label{sec:acronyms}

\begin{table*}[h]
\setlength{\tabcolsep}{2pt}
\captionof{table}{\label{tab:acronyms}\textcolor{black}{List of acronyms used.}}
\centering
\begin{tabular}{lll}
\toprule
\textbf{\textcolor{black}{Context}} & \textbf{\textcolor{black}{Acronym}} & \textbf{\textcolor{black}{Meaning}} \\
\midrule
\multirow{3}{*}{\textcolor{black}{Datasets}} & \textbf{\textcolor{black}{ADNI}} & \textcolor{black}{The Alzheimer's Disease Neuroimaging Initiative} \\
& \textbf{\textcolor{black}{HCP}} & \textcolor{black}{Human Connectome Project} \\
& \textbf{\textcolor{black}{PPMI}} & \textcolor{black}{The Parkinson's Progression Markers Initiative} \\
\midrule
\multirow{16}{*}{\textcolor{black}{Methods}} & \textbf{\textcolor{black}{CINTA}} & \textcolor{black}{Clustering in Tractography Using Autoencoders} \\
& \textbf{\textcolor{black}{CSD}} & \textcolor{black}{Constrained spherical deconvolution} \\
& \textbf{\textcolor{black}{DFC}} & \textcolor{black}{Deep Fiber Clustering} \\
& \textbf{\textcolor{black}{FIESTA}} & \textcolor{black}{FIbEr Segmentation in Tractography using Autoencoders} \\
& \textbf{\textcolor{black}{FINTA}} & \textcolor{black}{Filtering in Tractography Using Autoencoders} \\
& \textbf{\textcolor{black}{FINTA-m}} & \textcolor{black}{FINTA-multibundle} \\
& \textbf{\textcolor{black}{GESTA}} & \textcolor{black}{Generative Sampling in Bundle Tractography using Autoencoders} \\
& \textbf{\textcolor{black}{GESTA-gmm}} & \textcolor{black}{GESTA-Gaussian mixture model} \\
& \textbf{\textcolor{black}{PFT}} & \textcolor{black}{Particle Filtering Tractography} \\
& \textbf{\textcolor{black}{QBX}} & \textcolor{black}{QuickBundlesX} \\
& \textbf{\textcolor{black}{RB}} & \textcolor{black}{RecoBundles} \\
& \textbf{\textcolor{black}{RBX}} & \textcolor{black}{RecoBundlesX} \\
& \textbf{\textcolor{black}{SET}} & \textcolor{black}{Surface-Enhanced Tractography} \\
& \textbf{\textcolor{black}{TOM}} & \textcolor{black}{Tract Orientation Mapping} \\
& \textbf{\textcolor{black}{WMA}} & \textcolor{black}{White Matter Analysis} \\
& \textbf{\textcolor{black}{WMQL}} & \textcolor{black}{White Matter Query Language} \\
\midrule
\multirow{6}{*}{\textcolor{black}{Mathematics}} & \textbf{\textcolor{black}{ICC}} & \textcolor{black}{Intraclass correlation coefficient} \\
& \textbf{\textcolor{black}{k-NN}} & \textcolor{black}{k-nearest neighbors} \\
& \textbf{\textcolor{black}{MSE}} & \textcolor{black}{Mean squared error} \\
& \textbf{\textcolor{black}{PDF}} & \textcolor{black}{Probability distribution function} \\
& \textbf{\textcolor{black}{ROC}} & \textcolor{black}{Receiver operating characteristic} \\
& \textbf{\textcolor{black}{RS}} & \textcolor{black}{Rejection sampling} \\
\midrule
\multirow{12}{*}{\textcolor{black}{Miscellaneous}} & \textbf{\textcolor{black}{AD}} & \textcolor{black}{Alzheimer's disease} \\
& \textbf{\textcolor{black}{BOI}} & \textcolor{black}{Bundle of interest} \\
& \textbf{\textcolor{black}{dODF}} & \textcolor{black}{Diffusion orientation distribution function} \\
& \textbf{\textcolor{black}{fODF}} & \textcolor{black}{Fiber orientation distribution function} \\
& \textbf{\textcolor{black}{HC}} & \textcolor{black}{Healthy control} \\
& \textbf{\textcolor{black}{LOA}} & \textcolor{black}{Local orientation angle} \\
& \textbf{\textcolor{black}{MS}} & \textcolor{black}{Multiple sclerosis} \\
& \textbf{\textcolor{black}{PD}} & \textcolor{black}{Parkinson's disease} \\
& \textbf{\textcolor{black}{QC}} & \textcolor{black}{Quality control} \\
& \textbf{\textcolor{black}{ROI}} & \textcolor{black}{Region of interest} \\
& \textbf{\textcolor{black}{TR}} & \textcolor{black}{Test-retest} \\
& \textbf{\textcolor{black}{WBT}} & \textcolor{black}{Whole-brain tractography} \\
\midrule
\multirow{4}{*}{\textcolor{black}{Manuscript scope}} & \textbf{\textcolor{black}{IST}} & \textcolor{black}{Implausible streamlines from the TractoInferno validation set} \\
& \textbf{\textcolor{black}{PAWM}} & \textcolor{black}{Population average of white matter} \\
& \textbf{\textcolor{black}{SST}} & \textcolor{black}{Silver standard streamlines from the TractoInferno validation set} \\
& \textbf{\textcolor{black}{WBTT}} & \textcolor{black}{Whole-brain tractograms from the TractoInferno validation set} \\
\bottomrule
\end{tabular}
\end{table*}

\newpage
\subsection{\textcolor{black}{FIESTA's improvements over FINTA, CINTA, GESTA}}
\label{sec:improvements}
\textcolor{black}{As for FINTA, CINTA and GESTA, FIESTA is built upon autoencoders to process WBT. However, modifications were made to improve the practicality of FINTA, CINTA, and GESTA. The \textcolor{black}{FINTA-multibundle} module (Fig.~\ref{fig:ae_binta} (b)) integrates the notion of filtering and bundle segmentation as it is presented in FINTA and CINTA. However, FINTA works with an atlas of plausible streamlines to filter WBT based on a single whole-brain latent space threshold, whilst CINTA uses an atlas of plausible streamlines to segment bundles. CINTA, assumes that the input tractogram is already pre-filtered by another process (i.e., FINTA). To be more efficient, in our case, we merged the filtering and bundling tasks into a single process called \textcolor{black}{FINTA-multibundle}. Therefore, the module receives a WBT alongside an atlas of bundles and embeds those tractograms. In the latent space, with a $k$-NN approach, with $k=1$, each streamline inside the WBT to process is assigned to the closest streamline inside the atlas of bundles, therefore assigning a bundle class to each one of them. Finally, with a per-bundle distance threshold (instead of a single whole-brain threshold), we discard each streamline that is further than its corresponding distance threshold to its closest streamline in the atlas of bundle.}

\textcolor{black}{Regarding the \textcolor{black}{GESTA-gmm} module, we improved its efficiency by using a mixture ($N=11$) of multivariate Gaussians optimized with the expectation-maximization algorithm as a proposal distribution for the rejection sampling step instead of a single multivariate Gaussian. Also, we used the Silverman's rule of thumb ~\cite{silverman_density_1986} (instead of a cross-validation approach) to estimate automatically the kernel bandwidth of the Parzen estimator \cite{bishop_pattern_2006} for the latent space empirical PDF estimation.}

\newpage
\subsection{\textcolor{black}{Method's parameters}}
\label{sec:parameters}
\begin{table*}[h]
\setlength{\tabcolsep}{2pt}
\captionof{table}{\label{tab:methodsparameters}\textcolor{black}{List of principal parameters used}}
\centering
\begin{tabular}{lllc}
\toprule
\textbf{\textcolor{black}{Method}} & \textbf{\textcolor{black}{Parameter}} & \textbf{\textcolor{black}{Value}} & \textbf{\textcolor{black}{Time (minutes)}} \\
\midrule
\multirow{2}{*}{\textcolor{black}{\textit{TractSeg}}} & \textbf{\textcolor{black}{Number of fibers per bundle}} & \textcolor{black}{2000} & \multirow{2}{*}{\textcolor{black}{15}}\\
& \textbf{\textcolor{black}{Tracking algorithm}} & \textcolor{black}{Probabilistic} & \\

\midrule
\multirow{5}{*}{\textcolor{black}{RB}} & \textbf{\textcolor{black}{Atlas}} & \textcolor{black}{PAWM} & \multirow{5}{*}{\textcolor{black}{22}}\\
& \textbf{\textcolor{black}{WBT clustering threshold (mm)}} & \textcolor{black}{12} &\\
& \textbf{\textcolor{black}{Bundle clustering threshold (mm)}} & \textcolor{black}{4} &\\
& \textbf{\textcolor{black}{Far pruning distance (mm)}} & \textcolor{black}{18} &\\
& \textbf{\textcolor{black}{Local pruning distance (mm)}} & \textcolor{black}{6} &\\
\midrule
\multirow{5}{*}{\textcolor{black}{RBx}} & \textbf{\textcolor{black}{Atlas}} & \textcolor{black}{PAWM} &\multirow{5}{*}{\textcolor{black}{47}} \\
& \textbf{\textcolor{black}{WBT clustering threshold (mm)}} & \textcolor{black}{[15, 12]} &\\
& \textbf{\textcolor{black}{Bundle clustering threshold (mm)}} & \textcolor{black}{[3, 3.5, 4] } & \\
& \textbf{\textcolor{black}{Far pruning distance (mm)}} & \textcolor{black}{18} &\\
& \textbf{\textcolor{black}{Local pruning distance (mm)}} & \textcolor{black}{[5, 5.5, 6]} &\\
\midrule
\multirow{1}{*}{\textcolor{black}{WMA}} & \textbf{\textcolor{black}{Atlas}} & \textcolor{black}{WMA} & \multirow{1}{*}{\textcolor{black}{>180}}\\
\midrule
\multirow{1}{*}{\textcolor{black}{\textit{XTRACT}}} & \textbf{\textcolor{black}{Atlas}} & \textcolor{black}{\textit{XTRACT}} & \multirow{1}{*}{\textcolor{black}{>180}}\\ 
\midrule
\multirow{1}{*}{\textcolor{black}{FINTA-m}} & \textbf{\textcolor{black}{Atlas}} & \textcolor{black}{PAWM} & \multirow{1}{*}{\textcolor{black}{30}}\\
\midrule
\multirow{3}{*}{\textcolor{black}{GESTA-gmm}} & \textbf{\textcolor{black}{Atlas}} & \textcolor{black}{PAWM} & \multirow{3}{*}{\textcolor{black}{150}}\\ 
& \textbf{\textcolor{black}{Number of generated streamlines per bundle}} & \textcolor{black}{25000} &\\
& \textbf{\textcolor{black}{GMM components}} & \textcolor{black}{11} &\\
\bottomrule
\end{tabular}
\label{tab:method_params}
\end{table*}

\end{document}